\documentclass[runningheads]{llncs}
\usepackage{graphicx}
\usepackage{tikz}
\usepackage{comment}
\usepackage{amsmath,amssymb} % define this before the line numbering.
\usepackage{color}

\usepackage[accsupp]{axessibility}  % Improves PDF readability for those with disabilities.

\usepackage[width=122mm,left=12mm,paperwidth=146mm,height=193mm,top=12mm,paperheight=217mm]{geometry}

\usepackage{epsfig}
\usepackage{graphicx}
\usepackage{amsmath}
\usepackage{amssymb}
\usepackage{booktabs}
\usepackage{soul}
\usepackage{enumitem}
\usepackage{times}
\usepackage{lipsum}
\usepackage[normalem]{ulem}
\usepackage[export]{adjustbox}
\usepackage{graphbox}
\usepackage{dsfont}
\usepackage{subcaption}
\usepackage{comment}
\usepackage{arydshln}
\usepackage{float}
\usepackage{stfloats}
\usepackage{balance}
\usepackage[stable]{footmisc}
\usepackage{multirow}
\usepackage{changepage} 
\usepackage{etoc}
\usepackage[colorlinks=true, linkcolor=blue, urlcolor=blue, citecolor=blue, anchorcolor=blue]{hyperref}
\usepackage{bm}

\graphicspath{{figs/}}

\definecolor{stelios_colour}{RGB}{144, 238, 144}
\definecolor{light_red}{RGB}{255, 204, 204}
\definecolor{crimson}{rgb}{0.86, 0.08, 0.24}

\newif\ifcomment

\commentfalse

\ifcomment
\newcommand{\stelios}[1]{\sethlcolor{stelios_colour}\hl{[\textbf{Stelios:} #1]}}
\newcommand{\steve}[1]{\sethlcolor{cyan}\hl{[Steve: #1]}}
\newcommand{\nic}[1]{\sethlcolor{yellow}\hl{[Nic: #1]}}
\newcommand{\alex}[1]{\sethlcolor{orange}\hl{[Alex: #1]}}
\newcommand{\cut}[1]{\sethlcolor{light_red}\hl{[#1]}}
\newcommand{\blue}[1]{\textcolor{blue}{#1}}
\else
\newcommand{\stelios}[1]{}
\newcommand{\steve}[1]{}
\newcommand{\alex}[1]{}
\newcommand{\nic}[1]{}
\newcommand{\cut}[1]{}
\newcommand{\blue}[1]{\textcolor{black}{#1}}
\fi

 \usepackage{background}
 \backgroundsetup{angle=0,  
   scale=1, 
   color=black, 
   firstpage=true,
   position=current page.north, 
   hshift=0pt,
   vshift=-13pt,
   contents={\ifnum\value{page}=1 \small \textit{Preprint \hspace{9.5cm} Accepted at ECCV 2022} \else \fi}
 }

\begin{document}
% \renewcommand\thelinenumber{\color[rgb]{0.2,0.5,0.8}\normalfont\sffamily\scriptsize\arabic{linenumber}\color[rgb]{0,0,0}}
% \renewcommand\makeLineNumber {\hss\thelinenumber\ \hspace{6mm} \rlap{\hskip\textwidth\ \hspace{6.5mm}\thelinenumber}}
% \linenumbers
\pagestyle{headings}
\mainmatter

\def\ECCVSubNumber{7964}  % Insert your submission number here
\title{Multi-Exit Semantic Segmentation Networks} % Replace with your title

% INITIAL SUBMISSION 
%\titlerunning{ECCV-22 submission ID \ECCVSubNumber} 
%\authorrunning{ECCV-22 submission ID \ECCVSubNumber} 
%\author{Anonymous ECCV submission}
%\institute{Paper ID \ECCVSubNumber}
%\end{comment}
%******************

% CAMERA READY SUBMISSION
%\begin{comment}

\titlerunning{Multi-Exit Semantic Segmentation Networks}
% If the paper title is too long for the running head, you can set
% an abbreviated paper title here
%
\author{
Alexandros Kouris\inst{1}, 
Stylianos I. Venieris\inst{*,1}, 
Stefanos Laskaridis\inst{*,1}, 
Nicholas Lane\inst{1,2} 
}
\authorrunning{A. Kouris et al.}
% First names are abbreviated in the running head.
% If there are more than two authors, 'et al.' is used.
%
\institute{
$^1$Samsung AI Center, Cambridge \phantom{00} $^2$University of Cambridge  \phantom{00} $^*$Indicates equal contribution.\\
\email{\scriptsize\{a.kouris,s.venieris\}samsung.com, mail@stefanos.cc, nic.lane@samsung.com}
}

\maketitle

%%%%%%%%% ABSTRACT
\begin{abstract}
\noindent
Semantic segmentation arises as the backbone of many vision systems, spanning from self-driving cars and robot navigation to augmented reality and teleconferencing. Frequently operating under stringent latency constraints within a limited resource envelope, optimising for efficient execution becomes important. At the same time, the heterogeneous capabilities of the target platforms and the diverse constraints of different applications require the design and training of multiple target-specific segmentation models, leading to excessive maintenance costs. To this end, we propose a framework for converting state-of-the-art segmentation CNNs to Multi-Exit Semantic Segmentation (MESS) networks: specially trained models that employ parametrised early exits along their depth to \textit{i)}~dynamically save computation during inference on easier samples and \textit{ii)}~save training and maintenance cost by offering a post-training customisable speed-accuracy trade-off. Designing and training such networks naively can hurt performance. Thus, we propose a novel two-staged training scheme for multi-exit networks. Furthermore, the parametrisation of MESS enables co-optimising the number, placement and architecture of the attached segmentation heads along with the exit policy, upon deployment via exhaustive search in $<$1GPUh. This allows MESS to rapidly adapt to the device capabilities and application requirements for each target use-case, offering a train-once-deploy-everywhere solution. MESS variants achieve latency gains of up to 2.83$\times$ with the same accuracy, or 5.33 pp higher accuracy for the same computational budget, compared to the original backbone network. Lastly, MESS delivers orders of magnitude faster architectural customisation, compared to state-of-the-art techniques.
\end{abstract}
 
%%%%%%%%% BODY TEXT
\section{Introduction}
\label{sec:intro}
% Semantic Segmentation Emergence/Applications
Semantic segmentation constitutes a core machine vision task that has demonstrated tremendous advancement due to the emergence of deep learning~\cite{ghosh2019understanding}. By predicting dense (every-pixel) semantic labels for an image of arbitrary resolution, semantic segmentation forms one of the finest-grained visual scene understanding tasks, materialised as an enabling technology for myriad applications, including augmented reality~\cite{edge_ar2019mobicom,heimdall2020mobicom}, video conferencing~\cite{telepresence2020eccv,nvidia_video_conf}, navigation~\cite{Siam_2018_CVPR_Workshops,xu2017end}, and semantic mapping~\cite{mccormac2017semanticfusion}. 

% Call for efficient Inference
This wide adoption of segmentation models in consumer applications has pushed their deployment away from the cloud, towards resource-constrained edge devices~\cite{ai_benchmark_2019,heimdall2020mobicom} such as smartphones and home robots. With quality-of-service (QoS) and safety being of utmost importance when deploying such real-time systems, efficient and accurate segmentation becomes a core problem to solve.
% Call for efficient Adaptation
Additionally, device heterogeneity in the consumer ecosystem (\textit{e.g.}~co-existence of top-tier and low-cost smartphones) and the diverse constraints of different applications (\textit{e.g.}~\textit{30~fps} for AR/VR vs \textit{1~fps} for photo effects), call for segmentation models with variable latency-accuracy characteristics to be designed, trained and distributed to end devices, leading to high maintenance costs.

% Semantic Segmentation Workload/Latency Burden
State-of-the-art segmentation models, however, pose their own challenges to efficient deployment and adaptation, as their impressive accuracy comes at the cost of excessive computational and memory demands. Particularly, the every-pixel nature of the segmentation output calls for high-resolution feature maps to be preserved throughout the network (to avoid eradicating spatial information)~\cite{yu2017dilated}, while also maintaining a large receptive field on the output (to incorporate context and extract robust semantics)~\cite{peng2017large}, leading to inflated training and inference costs.

% Importance of Low latency Inference (and some related work directions)
Aiming to alleviate this latency burden for on-device inference~\cite{embench_2019}, recent work has focused on the design of lightweight segmentation models either manually~\cite{mehta2018espnet,zhao2018icnet} or through Neural Architecture Search~\cite{liu2019auto,nekrasov2019fast}. However, such methods typically involve huge search spaces and disallow the re-use of ImageNet \cite{deng2009imagenet} pre-trained classification backbones. This leads to long and non-reusable training cycles per model, which often differ for each target device, aggravating prohibitively the training and adaptation time.

Orthogonally, advances in early-exit DNNs offer complementary efficiency gains by adjusting the computation path at run time in an input-dependent manner, while natively providing a tunable speed-accuracy trade-off. However, these solutions~\cite{msdnet2018iclr,sdn2019icml,zhou2019edge} have mainly aimed at image classification so far, leaving challenges in segmentation, such as the design of lightweight exit architectures and exit policies, largely unaddressed. In fact, naively applying early-exiting on segmentation CNNs %can lead to degraded accuracy due to early-exit cross-talk during training, and potentially 
may not lead to any latency gains due to the inherently heavyweight architecture of segmentation heads, aggravated by the large incoming feature volume. For example, adding a single extra head on DeepLabV3~\cite{chen2017rethinking} leads to an overhead of up to 40\% of the original model's workload.

% Our one-size-fits-all Multi-Exit Segmentation Approach
In this work, we introduce a novel methodology for deriving and training Multi-Exit Semantic Segmentation (MESS) networks starting from existing CNNs and aiming for efficient and versatile on-device segmentation tailored to the platform and task at hand. MESS brings together architecture customisation and early-exit networks, through a novel training scheme and a compact and highly re-usable search space that allows post-training adaptation through exhaustive search in abridged time frames.

Specifically, MESS uses a given segmentation CNN as a backbone model, pre-trains it in an \textit{early-exit aware} manner (without loss of accuracy) and attaches numerous candidate early-exit architectures (\textit{i.e.}~segmentation heads) at different depths, offering predictions with varying workload-accuracy characteristics (Fig.~\ref{fig:cover}). Importantly, through targeted design choices, the \textit{number}, \textit{placement} and \textit{architecture} of exits remain configurable and can be co-optimised upon deployment, to adapt to different-capability devices and diverse application requirements, without the need of retraining, leading to a \textit{train-once, deploy-everywhere} paradigm. The main contributions of this work are: 
\begin{itemize}[leftmargin=.35cm]
    \setlength\itemsep{-0mm}
    \item The design of MESS networks, combining adaptive inference through early exiting with architecture customisation, to provide a fine-grain speed-accuracy trade-off, tailor-made for semantic segmentation. \textit{This enables efficient and adaptive segmentation based on the use-case requirements and the target device capabilities.}
    \item A two-stage scheme for training MESS networks, starting with an end-to-end exit-aware pre-training of the backbone that employs a novel exit-dropout loss which pushes the extraction of semantically strong features towards shallow layers of the network without compromising its final accuracy or committing to an exit configuration; followed by a frozen-backbone stage that jointly trains all candidate early-exit architectures through a novel selective distillation scheme. \textit{This mechanism boosts the accuracy of multi-exit networks and decouples training from the deployed MESS configuration, thus enabling rapid post-training adaptation of the architecture.}
    \item An input-dependent inference pipeline for MESS networks, employing a novel method for estimating the prediction confidence at each exit, used as exit policy, tailored for every-pixel outputs. \textit{This mechanism enables difficulty-based allocation of resources, by early-stopping for ``easy" inputs with corresponding performance gains.}
\end{itemize}

%%%%%%%%%%%%%%%%%%%%%%%
\vspace{-2.8mm}
\section{Related Work}
\label{sec:relwork}
\vspace{-1.2mm}
\noindent
\textbf{Efficient Segmentation.}~Semantic segmentation is rapidly evolving, since the emergence of the first CNN-based approaches~\cite{long2015fully,badrinarayanan2017segnet,noh2015learning,ronneberger2015u}. Recent advances have \mbox{focused} on optimising accuracy through stronger backbone CNNs~\cite{he2016deep,huang2017densely}, dilated convolutions
~\cite{yu2017dilated,chen2017deeplab}, multi-scale processing~\cite{yu2016mutli,zhao2017pyramid}, multi-path refinement~\cite{lin2017refinenet,ghiasi2016laplacian},  knowledge distillation \cite{liu2019structured} and adversarial training \cite{luc2016semantic}. To reduce the computational cost, the design of lightweight hand-crafted~\cite{mehta2018espnet,wu2019fastfcn,zhao2018icnet,yu2018bisenet} and more recently NAS-crafted~ \cite{nekrasov2019fast,liu2019auto,cardiacsegnas2020iccad} architectures has been explored. %with further efforts to compensate for the lost accuracy through . 
MESS is \textit{model-agnostic} and can follow the above advancements by being \textit{applied on top of existent CNN backbones} to achieve complementary gains by exploiting the orthogonal dimension of input-dependent early-exiting.
%Our methodology is able to follow the former advancements as the proposed framework can convert any segmentation model to a MESS network, while being orthogonal to the latter by exploiting the dimension of input-dependent inference that can provide complementary gains on top of efficient models.  

\vspace{1mm}\noindent
\textbf{Adaptive Inference.}~The key paradigm behind adaptive inference is to save computation on ``easy" samples and reduce the overall computation with minimal accuracy degradation~\cite{adaptive2017icml,figurnov2017spatially}.
% In this direction,
Existing methods can be taxonomised into: \textit{1) Dynamic Routing networks} selecting a different sequence of operations to run in an input-dependent manner by skipping layers~\cite{adaptive_layer2018eccv,wang2018skipnet,wu2018blockdrop} or channels~\cite{runtime_pruning2017neurips,gatingnns2019neurips,nestdnn2018mobicom,gao2019dynamic,wang2020dynamic}; and \textit{2) Multi-Exit Networks} forming a class of architectures with intermediate classifiers along their depth ~\cite{teerapittayanon2016branchynet,msdnet2018iclr,hapi2020iccad,deebert2020acl,xing2020early,yuan2019s2dnas}. With earlier exits running faster and deeper ones being more accurate, such networks provide varying accuracy-cost trade-offs. Existing work has mainly focused on image classification, proposing hand-crafted~\cite{msdnet2018iclr,scan2019neurips}, model-agnostic~\cite{teerapittayanon2016branchynet,sdn2019icml} and deployment-aware architectures~\cite{spinn2020mobicom,hapi2020iccad}. Yet, adopting these techniques in segmentation poses additional, still unexplored, challenges. %such as the parametrisation of effective light-weight exits and crafting of an exit-policy. 

\vspace{1mm}\noindent
\textbf{Multi-Exit Network Training.}~So far, the training of multi-exit models for classification can be categorised into:
\textit{1)~End-to-end} schemes jointly training the backbone and early exits~\cite{msdnet2018iclr,sdn2019icml,scan2019neurips},
%% The main benefit of this approach is the flexibility given to each exit to affect the extraction of semantics on the backbone leading to 
leading to increased accuracy in early exits, at the expense of often downgrading the accuracy deeper on or even causing divergence~\cite{msdnet2018iclr,ee_training2019iccv} due to early-exit ``cross-talk"; and \textit{2)~Frozen-backbone} methods which firstly train the backbone until convergence and subsequently attach and train intermediate exits individually \cite{sdn2019icml,hapi2020iccad}. 
This decoupling of the backbone from the exits allows for faster training of the exits, at the expense of an accuracy penalty due to fewer degrees of freedom in parameter tuning. 
Orthogonally, self-distillation methods have been proposed in the literature~\cite{ee_training2019iccv,phuong2019distillation,zhang2019be,reda2020mm,persephonee2021hotmobile} to further improve the accuracy of early exits by treating them as students of the last exit. 
% A complementary approach that aims to further improve the early exits' accuracy involves \textit{knowledge distillation} between exits, studied in classification~\cite{ee_training2019iccv,phuong2019distillation,zhang2019be} and domain adaptation \cite{reda2020mm,persephonee2021hotmobile} tasks. Such schemes employ self-distillation by treating the last exit as the teacher and the intermediate classifiers as the students, without priors about the ground truth.
In this work, we propose a fused two-stage training scheme, backed by self-distillation with information filtering, that enables exit-aware pre-training and full customisation potential without affecting the final exit's accuracy. 
% In this work, we introduce a novel two-stage training scheme for MESS networks, consisting of an exit-aware backbone training step that pushes the extraction of semantically ``strong" features early in the network, followed by a frozen-backbone step for fully training the early exits without affecting the final exit's accuracy.

\noindent
\textbf{Adaptive Segmentation Networks.}
Recently, initial efforts on adaptive segmentation have emerged. Li~\textit{et al.}~\cite{li2020learning} combined NAS with a trainable dynamic routing mechanism that generates data-dependent processing paths at run time. NAS approaches, however, compose enormous search spaces with minimum re-use between instances, 
%while their formulation does not allow to exploit ImageNet pre-trained classification models, 
leading to soaring training times. Furthermore, by incorporating the computation cost to the loss function, this approach is unable to customise the model to meet varying speed-accuracy characteristics %for diverse application requirements or across heterogeneous devices 
without retraining, leading also to inflated adaptation cost. Closer to our work, Layer Cascade (LC)~\cite{li2017not} studies \textit{early-stopping} for segmentation. LC treats segmentation as a vast group of independent classification tasks, where \textit{each pixel} propagates to the next exit only if the latest prediction does not surpass a confidence threshold. Nonetheless, due to different \textit{per-pixel} paths, this scheme leads to heavily unstructured computations, for which existing BLAS libraries cannot achieve realistic speedups~\cite{balancedsparse2019aaai}. LC also constitutes a manually-crafted model, tied to a specific backbone architecture, and non-customisable to the target device's capabilities. 

\vspace{1mm}
MESS networks bring together benefits of all the above worlds. Our framework supports model customisation within a compact search space of early-exit architectures tailor-made for semantic segmentation, while preserving the ability to re-use pre-trained backbones cutting down training time. Additionally, MESS networks push the limits of efficient inference by incorporating \textit{image-level} confidence-based early exiting, through a novel exit policy that addresses the unique challenges of dense segmentation predictions. Simultaneously, the proposed two-stage training scheme combines end-to-end and frozen-backbone training approaches, boosting the accuracy of shallow exits without compromising deeper ones. Finally, design choices allow us to decouple MESS training from the deployment configuration, enabling exhaustive search to be rapidly performed post-training, in order to customise the architectural configuration for different devices or application-specific requirements, without any parameter fine-tuning.

%%%%%%%%%%%%%%%%%%%%%%%
\section{MESS Networks Overview}
\label{sec:methodology_overview}
\begin{figure*}[t]
   \begin{minipage}[t]{0.5\textwidth}
      \centering
	  \includegraphics[trim =24mm 40mm 42mm 45mm, clip,width=1.0\columnwidth]{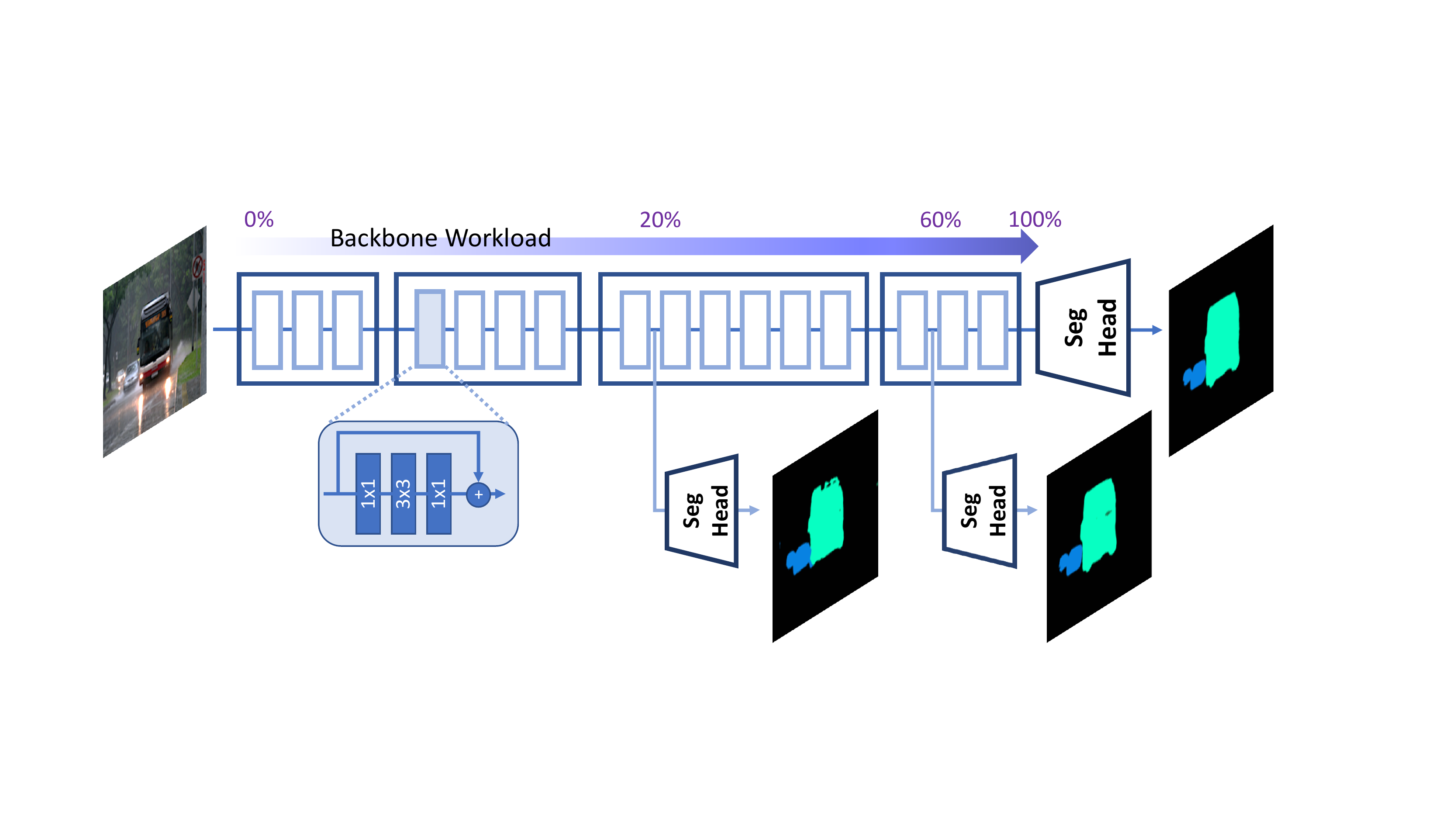}
	  %\vspace{-6mm}
	  \captionof{figure}{\scriptsize Multi-Exit Semantic Segmentation network instance. \steve{Bring these figures up}}
	   \label{fig:cover}
    \end{minipage}
\hfill  
   \begin{minipage}[t]{0.47\textwidth}
       \centering
	   \includegraphics[trim =101mm 79mm 103mm 68mm, clip,width=1.0\columnwidth]{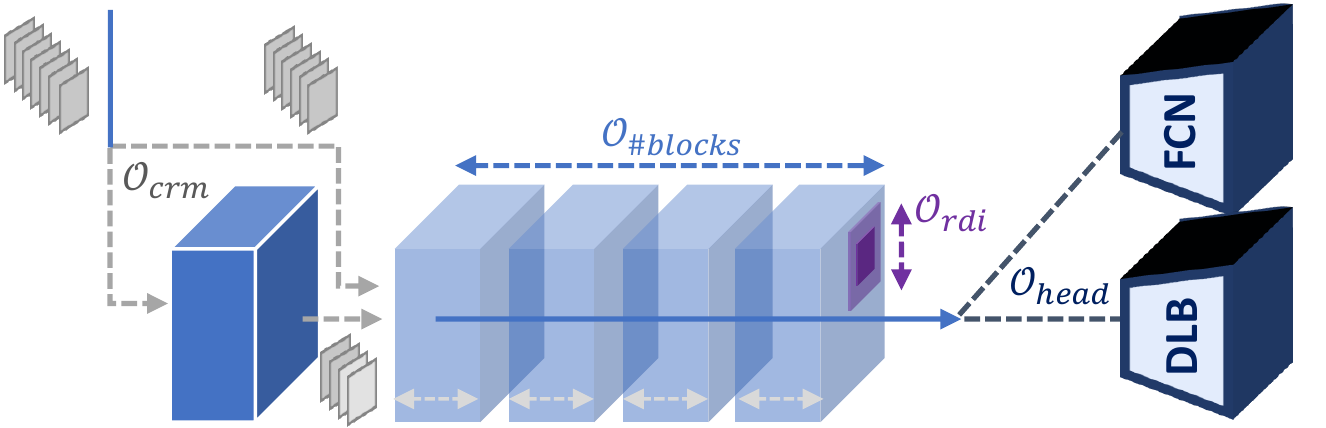}
	   %\vspace{-6mm}
	   \captionof{figure}{\scriptsize Parametrisation of segmentation head architecture.\steve{Refer to where the parameters are defined.}}
	   \label{fig:exit_arch}
       \end{minipage}
    %\vspace{-3mm}
\end{figure*}

To enable efficient segmentation, the MESS \textit{framework} employs a target-specific configuration search to obtain a \textit{multi-exit} segmentation network optimised for the platform and task at hand. We call the resulting model a MESS network, with an example depicted in Fig.~\ref{fig:cover}.  
Constructing a MESS network involves three stages: \textit{i)} starting from a backbone segmentation CNN, we identify several candidate \textit{exit points} along its depth (Sec.~\ref{sec:backbone}), and \textit{attach} to each of them multiple \textit{early exits} (\textit{i.e.}~segmentation heads) of varying architectural configurations (Sec.~\ref{sec:exitarch}), %, accuracy and computational overhead, 
leading to a newly defined \textit{overprovisioned} network; \textit{ii)} \textit{training} all candidate exits together with the backbone through a novel two-stage scheme (Sec.~\ref{sec:training});
and \textit{iii)}~\textit{tailoring} the overprovisioned network post-training to extract a MESS \textit{instance}, comprising the backbone and a subset of the available exits, considering user-defined constraints and optimisation objectives  (Sec.~\ref{sec:deployment}). Our framework supports various inference settings, ranging from extracting efficient target-specific submodels (meeting accuracy/speed constraints) to progressive refinement of the segmentation prediction and confidence-based exiting (Sec.~\ref{sec:policy}). Across all settings, MESS networks save computation by circumventing deeper parts of the network. The next two sections follow the flow of the proposed framework.

\section{MESS Networks Design \& Training}
\label{sec:methodology_training}
In this section, we go through the design choices that shape MESS networks, their early-exit architectural configuration and training process. This yields an overprovisioned network, ready to be customised for the target application and device at hand. 

\subsection{Backbone Initialisation \& Exit Placement}
\label{sec:backbone}
Initially, a backbone segmentation CNN is provided. Typically, such models aim to preserve large receptive field on the output, while preventing loss of spatial information (\textit{e.g.}~by replacing traditional pooling operations with dilated convolutions \cite{yu2017dilated}). As a result, and combined with the increased number of channels integrated, deeper layers demonstrate significantly larger feature volumes, leading to an unbalanced distribution of computational demands across the network (Fig. \ref{fig:cover}). This motivates the adoption of early-exiting during inference as a means of improving processing speed.

%Typical semantic segmentation CNNs try to prevent the loss of spatial information that inherently occurs in classification, without reducing the receptive field on the output pixels. For example, Dilated Residual Networks~\cite{yu2017dilated} allow up to 8$\times$ spatial reduction in the feature maps, and replace any further traditional downsampling with a doubling in dilation rate in convolution operations. We adopt a similar assumption for the backbones used to generate MESS networks.

%This approach, however, increases the feature resolution in deeper layers, which usually integrate a higher number of channels. As a result, typical CNN architectures for segmentation contain workload-heavier layers deeper on, leading to an unbalanced distribution of computational demands and an increase in the overall workload. This fact further motivates the need for early-exiting in order to eliminate unnecessary computation and improve the processing speed. 

As a first step, the provided backbone is profiled in terms of per-layer  workload (FLOPs). Based on the results of this analysis, $N$ \textit{candidate exit points} are identified following an approximately equidistant workload distribution (every $1/N$-th of the total backbone's FLOPs). For simplicity, exit points are restricted to be at the output of individual network blocks\footnote{\textit{e.g.}~Dilated Residual Blocks for ResNet-based \cite{he2016deep} backbones, Inverted Residual Blocks for MobileNet-based \cite{sandler2018mobilenetv2} backbones etc.} $b_k$. Although some of these exit points may subsequently be dropped during MESS configuration search, this placement currently maximises the distance between subsequent exits, improving the efficiency of our search. An example of the described analysis on a DRN-50 backbone \cite{yu2017dilated} is presented in Fig. \ref{fig:work_layer}.

\subsection{Early-Exit Architecture Search Space Design}
\label{sec:exitarch}
Early-exiting in segmentation CNNs faces the challenge of: \textit{i)}~\textit{enlarged feature volumes} of segmentation models, leading to inflated computation cost for the early-exit heads, \textit{ii)}~\textit{limited receptive field} and \textit{iii)}~\textit{weak semantics} in shallow exits. MESS addresses these challenges in a two-fold manner: \textit{i)}~by pushing the extraction of semantically strong features to shallower layers of the backbone during training (Sec.~\ref{sec:training}) and \textit{ii)}~by introducing a configuration space tailored-made for segmentation head architectures:
%, allowing each exit to adopt a tailored architecture based on its position in the backbone.

\begin{adjustwidth}{2mm}{}
\noindent
\textbf{\textit{1) Channel Reduction Module (CRM):}}  
%A main differentiating challenge of early-exiting in segmentation, compared to classification, is the significantly higher workload of segmentation heads, stemming from the enlarged input feature volume being processed. 
To reduce the computational overhead of each exit without compromising the spatial resolution of the feature volume that is particularly important for accuracy, we optionally include a 1$\times$1 convolutional layer (CRM) that reduces the number of channels fed to the segmentation head by a tunable factor.
% we focus our optimisation efforts on the channel dimension. In this direction, the proposed configuration space includes the optional addition of a lightweight CRM, comprising a 1$\times$1 convolutional layer that rapidly reduces the number of channels fed to the segmentation head by a tunable factor.
% , alleviating partly its computational burden.

\noindent
\textbf{\textit{2) Extra Trainable Blocks:}}
To address the weak semantics of shallow exits, while avoiding an unnecessary surge in the computational overhead of deeper exits, we allow incorporating a configurable number of additional convolutional blocks in each exit's architecture.
% , by exposing this option as a \textit{configurable} parameter in our search.
These layers are tactically appended \textit{after} the CRM to take advantage of the computational efficiency of its reduced feature-volume width.

%Classification-centric approaches address feature extraction limitations of early classifiers by incorporating additional layers in each exit~\cite{teerapittayanon2016branchynet}. Again, due to the enlarged volume of feature maps in segmentation networks, naively introducing vanilla-sized layers may result to a surge in the exit's workload overhead, defeating the purpose of early-exiting. In MESS networks, we expose this as a \textit{configurable} option that can be employed to remedy weak semantics in shallow exits, while we carefully append such layers \textit{after} the CRM in order to take advantage of the computational efficiency of the reduced feature-volume width.

\noindent
\textbf{\textit{3) Rapid Dilation Increase (RDI):}} To address the limited receptive field of shallow exits, apart from the addition of trainable blocks, we optionally allow the dilation rate employed in the exit layers to be rapidly increased, doubling in each block.

\noindent
\textbf{\textit{4) Head:}} MESS currently supports two types of output segmentation blocks from the literature, positioned at the end of each exit:
\textit{i)}~Fully Convolutional Network-based Head (\textit{FCN-Head})~\cite{long2015fully} and \textit{ii)}~DeepLabV3-based Head (\textit{DLB-Head})~\cite{chen2017rethinking,chen2018encoder}. 
%The former provides a simple and effective mechanism for upsampling the feature volume through de-convolution \cite{Noh_2015_ICCV} and predicting a per-pixel probability distribution across all candidate classes. The latter incorporates Atrous Spatial Pyramid Pooling (ASPP) comprising parallel convolutions of different dilation rates in order to incorporate multi-scale contextual information to its predictions.
\end{adjustwidth}
\vspace{1mm}\noindent
Overall, the configuration space for each exit architecture (Fig. \ref{fig:exit_arch}) is shaped as:
\begin{enumerate}
    \setlength\itemsep{-0mm}
    \footnotesize
    \item  Channel Reduction Module: $\mathcal{O}_{\text{crm}}=$ \{$/1$, $/2$, $/4$, $/8$\} $\mapsto$ \{0, 1, 2, 3\}
    \item Extra Trainable Blocks: $\mathcal{O}_{\text{\#blocks}} =$ \{0, 1, 2, 3\}
    \item \footnotesize Rapid Dilation Increase: $\mathcal{O}_{\text{rdi}} =$ \{\textit{False}, \textit{True}\} $\mapsto$ \{0, 1\}
    \item \footnotesize Segmentation Head: $\mathcal{O}_{\text{head}} =$ \{\textit{FCN-Head}, \textit{DLB-Head }\} $\mapsto$ \{0, 1\}
\end{enumerate}

Expecting that varying exit-point depths favour different architectural configurations (\textit{e.g.}~channel-rich for deeper exits and layer-multitudinous for shallower), MESS allows each early exit to adopt a tailored architecture based on its position in the backbone. Formally, we represent the configuration space for the i-th exit's architecture as: %\phantom{0}
%$\mathcal{S}_{\text{exit}}^i =
%\mathcal{O}_{\text{crm}} \times \mathcal{O}_{\text{\#blocks}} \times
%\mathcal{O}_{\text{rdi}} \times \mathcal{O}_{\text{head}}  \inlineeqno$ %\phantom{0}
\begin{equation}
%\resizebox{0.88\linewidth}{!}{$
\footnotesize
\mathcal{S}_{\text{exit}}^i =
\mathcal{O}_{\text{crm}} \times \mathcal{O}_{\text{\#blocks}} \times
\mathcal{O}_{\text{rdi}} \times \mathcal{O}_{\text{head}}
\label{eq:layer_space}
\end{equation}
where $i \in\{1,2,...,N\}$ and $\mathcal{O}_{\text{crm}}$, $\mathcal{O}_{\text{\#blocks}}$, $\mathcal{O}_{\text{rdi}}$ and $\mathcal{O}_{\text{head}}$,
are the sets of available \textit{options} for the CRM, number of trainable blocks, RDI and exit head, respectively.

%MESS allows for distinct architectures to be adopted at each exit.
% 
% \steve{Maybe it would be beneficial to list these tuning knobs in the beginning of the subsection too or put bold paragraph titles to make the contributions more clear to the eye of the reviewer. Can be tied to Figure 3.}

% Steve: I removed this
% Notably, the majority of related work employ a uniform architecture for all exits for the sake of simplicity~\cite{msdnet2018iclr,sdn2019icml,scan2019neurips,hapi2020iccad}. However, as demonstrated in Sec.~\ref{sec:experiments}, different exit depths pose their own challenges, with shallow exits benefiting the most from numerous lightweight layers, whereas deeper exits favour channel-rich exit architectures. Our framework favours customisation, enabling the efficient search for a model with tailored architecture at each exit, through a two-stage training scheme (Sec.~\ref{sec:training}). % introduce below.
% introduced in the following section. 

%With shallow exits benefiting the most from numerous lightweight layers, whereas deeper exits favouring channel-rich exit architectures, MESS allows for 

\subsection{Training Scheme}
\label{sec:training}

%Having the network architecture set, we move to the training methodology of our framework, comprising a two-stage pipeline enhanced with positive filtering distillation.

\subsubsection{Two-Stage MESS Training.}
\label{sec:pre-training}

% \stelios{The subsection title refers only to the 1st stage.} \alex{'Two-stage MESS Training' ?} 
% \steve{Here I would flip the narrative to be more like the reader participating in the design. This means that I would first say the hypothesis, and then provide the design implementing the hypothesis, rather than saying this is the design and this is what it accomplishes.}
As aforementioned, early-exit networks are typically either trained \textit{end-to-end} or in a \textit{frozen-backbone} manner~\cite{adaptive_dnns2021emdl}. However, both can lead to suboptimal accuracy results in the final or the early exits. For this reason, we combine the best of both worlds by proposing a novel two-stage training scheme.

\vspace{1mm} \noindent
\textbf{\textit{Stage 1 (end-to-end):}}~In the exit-aware pre-training stage, we aim to fully train the backbone network that will be shared across all candidate exits, specially preparing it for early-exiting by pushing the extraction of semantically strong features at shallow layers, without committing to any particular exit configuration. To achieve this, vanilla FCN-Heads are attached to all candidate exit points, generating an intermediate multi-exit model. This network is trained end-to-end, updating the weights of the backbone and a \textit{single} early exit at each iteration, with the remainder of the exits being dropped-out in a round-robin fashion (Eq.~(\ref{eq:drop}), referred to as \textit{exit-dropout loss}). As a result, cross-talk between exits is minimised allowing the final head to reach its full potential, while the backbone remains exposed to gradients from shallower exits. Formally, we denote the segmentation predictions after softmax for each early exit by $\small \boldsymbol{y}_i \in [0,1]^{R\times C \times M}$ where $R$ and $C$ are the output's number of rows and columns, respectively, and $M$ the number of classes. Given the ground truth labels \mbox{$\small \hat{\boldsymbol{y}} \in \{0,1,...,M$-$1\}^{R\times C}$}, the loss function for the proposed exit-aware pre-training stage is formulated as:
\begin{equation}
%\resizebox{0.88\linewidth}{!}{$
\footnotesize
    \mathcal{L}_{\text{pretrain}}^{\text{batch}^{(j)}} =  \sum_{i=1}^{N-1}{\mathds{1}(j\ \text{mod}\ i=0) \cdot \mathcal{L}_{\text{CE}}(\boldsymbol{y}_i,\hat{\boldsymbol{y}}) +  \mathcal{L}_{\text{CE}}(\boldsymbol{y}_N,\hat{\boldsymbol{y}}) }
    \label{eq:drop}
\end{equation}
where $\mathds{1}(\cdot)$ is the indicator function and $\mathcal{L}_{\text{CE}}$ the cross entropy. Although after this stage the early exits are not fully trained, their contribution to the loss \textit{guides the backbone towards learning stronger representations throughout, consequently aiding early-exiting.}
%pushes the backbone to extract semantically stronger features even at shallow layers. 

\vspace{1mm} \noindent
\textbf{\textit{Stage 2 (frozen-backbone):}}~At this stage, the backbone and final exit are kept frozen (\textit{i.e.}~weights are not updated). The \textit{MESS overprovisioned network} is formed by attaching \textit{all} candidate early-exit architectures of the proposed configuration space $\mathcal{S}_{\text{exit}}^i$  (Sec.~\ref{sec:exitarch}) across \textit{all} candidate exit points $i \in \{1,2,...,N\}$ (Sec.~\ref{sec:backbone}) and training them jointly. Importantly, keeping the backbone unchanged during this stage allows different exit architectures to be: \textit{i)} attached and trained \textit{simultaneously} even to the same candidate exit point \textit{without interfering} with each other, or with the backbone \textit{ii)} trained at significantly reduced cost than the end-to-end approach, while taking advantage of the strong semantics extracted by the backbone due to its early-aware pre-training and \textit{iii)} interchanged at deployment time on top of the shared backbone in a \textit{plug-and-play} manner (without re-training), offering enormous flexibility for customisation (Sec.~\ref{sec:deployment}).

\vspace{-1mm}
\subsubsection{Positive Filtering Distillation (PFD).}
\label{sec:distillation}
In the second stage of our training process, we also exploit the joint potential of knowledge distillation and early-exit networks. 

In prior self-distillation works for multi-exit networks, the backbone's final output is used as the teacher for earlier classifiers~\cite{zhang2019be}, whose loss function typically combines ground-truth and distillation terms~\cite{phuong2019distillation,luan2019msd}. To further exploit what information is backpropagated to the shallow exits, given the pre-trained final exit and taking advantage of the multitude of information available in segmentation predictions due to their dense structure, we propose \textit{Positive Filtering Distillation} (PFD). This technique selectively controls the flow of information of the high-entropy ground-truth reference to earlier exits using only signals from ``easier pixels", \textit{i.e.}~pixels about which the last exit could yield a correct prediction, while filtering out gradients from more difficult or ambiguous pixels. Our hypothesis is that early-exit heads, having limited learning capacity, can become stronger by only incorporating signals of less ambiguous pixels from the last exit, avoiding noisy gradients and the confusion of trying to mimic contradicting references.

Formally, we express the i-th exit's tensor of predicted classes for each pixel $\small \boldsymbol{p} = (r,c)$ with $\small r$$\in$$ [1,R]$ and $\small c$$\in$$ [1,C]$ as $ \hat{\boldsymbol{y}}_i$$\in$$\{0,1,...,M$-$1\}^{R\times C}$ where $\small (\hat{\boldsymbol{y}}_i)_{\boldsymbol{p}} =\arg \max{} (\boldsymbol{y}_i)_{\boldsymbol{p}}$ in $\{0,1,...,M$-$1\}$. Given the corresponding output of the final exit $\hat{\boldsymbol{y}}_N$, the ground-truth labels \mbox{\small $\hat{\boldsymbol{y}}$$\in$$\{0,1,...,M$-$1\}^{R\times C}$} and a hyperparameter $\alpha$, we employ the following loss during the frozen-backbone stage of our training scheme, where $\mathcal{L}_{\text{KL}}$ is KL-divergence: 
\begin{equation}
    \footnotesize
    \label{eq:loss}
     %\mathcal{L}_{\text{PFD}} = \sum_{i=1}^{N} \alpha \cdot \mathcal{L}_{\text{KL}}(\boldsymbol{y}_i , \boldsymbol{y}_N) + (1-\alpha) \cdot \mathds{1}(\hat{\boldsymbol{y}}_N=\hat{\boldsymbol{y}}) \mathcal{L}_{\text{CE}}(\boldsymbol{y}_i,\hat{\boldsymbol{y}} )
     \mathcal{L}_{\text{PFD}} = \sum_{i=1}^{N} \alpha \cdot \mathds{1}(\hat{\boldsymbol{y}}_N=\hat{\boldsymbol{y}}) \mathcal{L}_{\text{CE}}(\boldsymbol{y}_i,\hat{\boldsymbol{y}} ) + (1-\alpha) \cdot \mathcal{L}_{\text{KL}}(\boldsymbol{y}_i , \boldsymbol{y}_N)
\end{equation}

\section{MESS Networks Deployment \& Inference}
%\vspace{-1mm}
\label{sec:methodology_inference}
Having designed and trained the \textit{overprovisioned} model, here we discuss its customisation to the task- and target-specific deployment for inference. This involves configuring the MESS \textit{instance} architecture via post-training search and crafting the exit policy. 

\subsection{Deployment-time Parametrisation}
\label{sec:deployment}
Post-training of the overprovisioned network (comprising \textit{all} candidate exit architectures), MESS instances (comprising a \textit{subset} of the trained exits) can be derived, reflecting on the capabilities of the target device, the required accuracy or latency of the use-case and the intricacy of the inputs.

%Having trained the overprovisioned network comprising all candidate exit architectures, MESS instances can be derived for the use-case at hand by exhaustive architectural search, reflecting on the capabilities of the target device, the intricacy of the inputs and the required accuracy or latency. 

\vspace{1mm} \noindent
\textbf{Inference Settings.}
To satisfy performance needs under each device and application-specific constraints, MESS networks support different inference settings:

\begin{figure*}[t]
   \begin{minipage}[t]{0.43\textwidth}
   \vspace{-3mm}
        \centering
    	\includegraphics[trim={0mm 3mm 0mm 0mm}, clip,width=1\columnwidth]{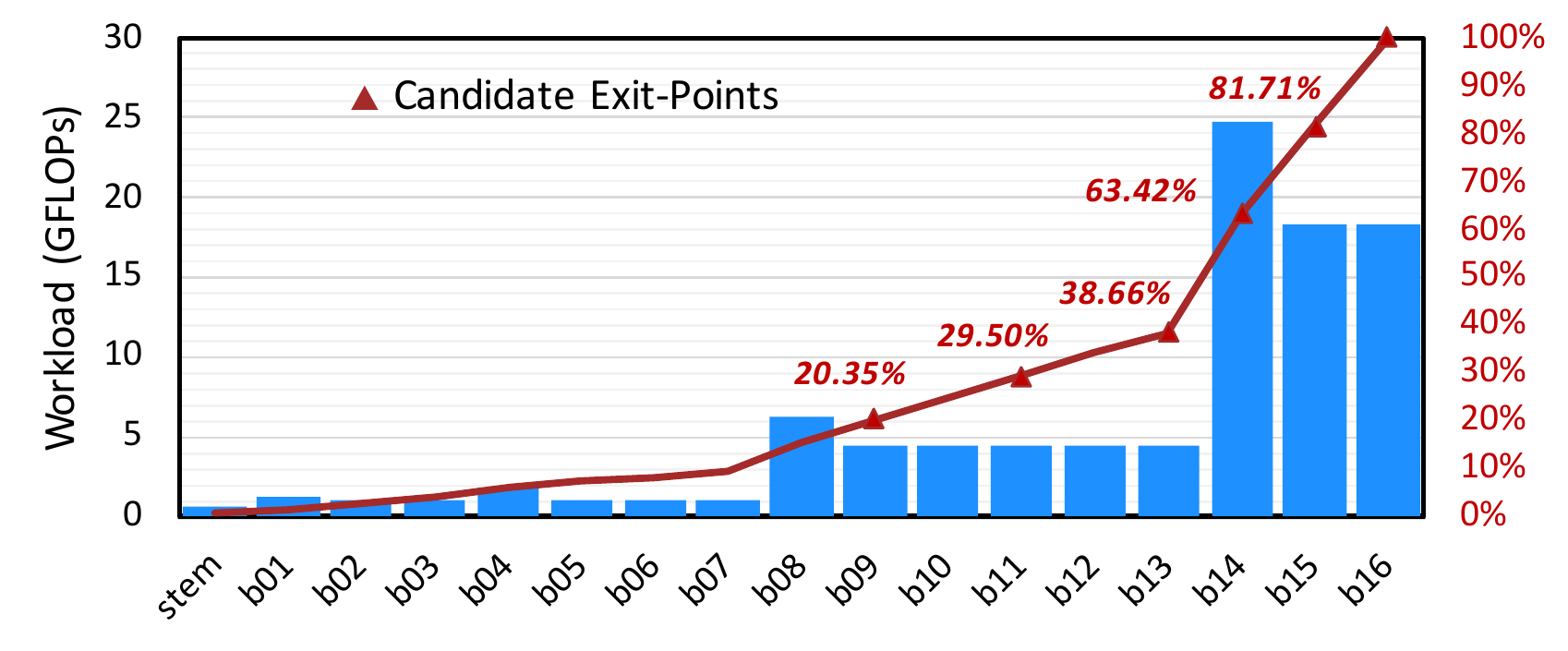}
    	\vspace{-8mm}
    	\caption{\scriptsize Workload breakdown analysis and exit points identification on a DRN-50 \cite{yu2017dilated} backbone ($N$=6).}
    	\label{fig:work_layer}
   \end{minipage}
\hfill  
   \begin{minipage}[t]{0.54\textwidth}
   \vspace{-4mm}
       \captionof{table}{\scriptsize Cost functions for different inference settings. $b_i$ is the i-th block in the backbone; $K_n$ is the block ordinal of the n-th exit point; $\mathcal{S^*}_{\text{exit}}^n \in \mathcal{S}_{\text{exit}}^n$ is the selected architecture for the n-th exit; $p_n$ is the percentage of samples propagated to the n-th exit.}
       \vspace{-2mm}
        \resizebox{1\columnwidth}{!}{
        \begin{tabular}{ cc}
         \hline
          \textbf{Inference} & \textbf{$cost(s)$} \\
          \hline
            Final-Only & $cost(b_{1:K_N}) + cost(\mathcal{S^*}_{\text{exit}}^N) $ \\
            Budgeted   & $cost(b_{1:K_n}) + cost(\mathcal{S^*}_{\text{exit}}^n)$ , $n \le N $  \\
            Anytime    & $cost(b_{1:K_N}) + \sum_{n=1}^{N} cost(\mathcal{S^*}_{\text{exit}}^n)$  \\
            Input-Dep. & $\sum_{n=1}^{N} p_{n-1} \cdot ( cost(b_{K_{n-1}:K_n}) +  cost(\mathcal{S^*}_{\text{exit}}^n) ) $ , $K_0$=0 , $p_0$=1 \\
            \hline
            \label{tab:inference_cost} 
        \end{tabular}
        }
   \end{minipage}
   \vspace{-3mm}
\end{figure*}

%\steve{Make this and the configuration space lists slightly indented. It's very flat atm.}
% At inference time, MESS networks can follow different inference settings:
\begin{adjustwidth}{2mm}{}
\noindent
\textbf{\textit{1)}~\textit{Budgeted Inference}:} 
in which workload-lighter static submodels, up to a (single) specific exit, are extracted to enable deployment on heterogeneous target platforms. \\ 
\textbf{\textit{2)}~\textit{Anytime Inference}:} in which every sample goes through multiple exits sequentially, initially providing a rapid approximation of the output and progressively refining it through a series of deeper exits until a deadline is met. \\ 
\textbf{\textit{3)}~\textit{Input-dependent Inference}:} where inputs also go through exits sequentially, but each sample dynamically adjusts its path (\textit{i.e.}~finalises its output at a different depth) according to its difficulty, as captured by the confidence of each exit's prediction. % (Sec.~\ref{sec:policy}).
\end{adjustwidth}
% 
% We envision further use-cases for MESS networks, including partitioning the model for cloud-device execution, preserving the ability to obtain an approximation of the final prediction relying solely on the on-board computational resources (to address networking issues)~\cite{spinn2020mobicom}, and incorporating ``specialist" exits to handle segmentation tasks focused on particular classes, input distributions or domains \cite{jiang2020resource}.

\vspace{1mm}\noindent
\textbf{Configuration Search.}
Our framework tailors MESS networks for each of the above settings considering the target use-case,  by searching the configuration space post-training. Contrary to most works in multi-exit classification models~\cite{msdnet2018iclr,sdn2019icml,scan2019neurips,hapi2020iccad}, which employ a uniform architecture across all exits for the sake of simplicity, MESS favours flexibility allowing for per-exit architectural customisation. This is enabled by our overprovisioned training scheme (Sec. \ref{sec:pre-training}), allowing all trained exits to be interchangeably attached to the same backbone for inference, offering rapid validation of candidate choices that significantly accelerates the search for a tailored design.

%This is enabled by the rapid validation of candidate choices, as they can be interchangeably attached to the same backbone for inference. Thus, the search for the optimal design is significantly accelerated, owning to our overprovisioned training scheme (Sec. \ref{sec:pre-training}).

% The proposed 
% Our framework tailors 
% MESS networks for each of the above settings considering the target use-case, through an \textit{exhaustive search} of the configuration space. In contrast to the majority of related work in multi-exit classification models that employ a uniform architecture across all exits for the sake of simplicity~\cite{msdnet2018iclr,sdn2019icml,scan2019neurips,hapi2020iccad},MESS favours customisation enabling the efficient search for a model with tailored architecture at each exit. This is enabled by the proposed two-stage training scheme (Sec. \ref{sec:pre-training}),  allowing all trained exits to be interchangeably attached to the same backbone for inference. % as \textit{plug-and-play} components.

%Concretely, we search for the number, placement and architecture of early exits, along with the exit policy for the \textit{input-dependent inference} case.

%\subsubsection{Number, Placement \& Configuration of Exits}
%\vspace{-0.5em}
%\label{sec:search}
%\noindent\textbf{Number, Placement \& Configuration of Exits.}
The proposed method contemplates all trained exit architectures and exhaustively creates different configurations, trading for example a workload-heavier shallow exit with a more lightweight deeper exit. The search strategy considers the target \textit{inference setting}, along with user-specified requirements in \textit{workload}, \textit{latency} and \textit{accuracy}\footnote{Evaluated in a held-out \textit{Calibration Set} during search (equally sized to the target \textit{Test Set}).}, which can be expressed as a combination of hard constraints and optimisation objectives. As a result, the \textit{number} and \textit{placement} of exits and the \textit{architecture} of each individual exit of the resulting MESS instance are jointly optimised (along with the \textit{exit policy}, discussed in Sec. \ref{sec:policy}, for the input-dependent inference case).
%are co-optimised in a joint exploration.

% We formally capture the number and position of exits by means of a positioning vector $\boldsymbol{p}_{\text{exit}} \in \{0,1 \}^N$ with the i-th element set to 1 if an exit is placed after layer group $i$. Furthermore, we define the positioning option set $\mathcal{O}_{\text{pos}}$ containing all $2^N$ candidate combinations of exits.
% \alex{An alternative way to formulate this (which may be more intuitive imo) is to add a 'None' option for each of the exit positions, instead of $O_{pos}$. So $\mathcal{S} = (\mathcal{S}_{\text{exit}}^1+1) \times (\mathcal{S}_{\text{exit}}^2+1)  \times ... $} \stelios{I see the point. I can't decide. @Stefanos?}
Given the exit-architecture search space $\mathcal{S}_{\text{exit}}^i$ (Eq.~\ref{eq:layer_space}), we define the overall configuration space of a MESS network as:
\begin{equation}
\footnotesize
\mathcal{S} = (\mathcal{S}_{\text{exit}}^1+1) \times (\mathcal{S}_{\text{exit}}^2+1)  \times ... \times (\mathcal{S}_{\text{exit}}^N+1)
\label{eq:mess_space}
\end{equation}
where the extra term accounts for a ``\textit{None}" option for each of the exit positions.
Under this formulation, the framework can minimise $workload/latency$, formally expressed as $cost$ for each setting in Table \ref{tab:inference_cost}, given an $accuracy$ constraint $th_{\text{acc}}$:
\begin{equation}
    \small
    s^{\star} = \arg\min_{s \in \mathcal{S}} \ \{ \text{cost}(s) \ | \ \text{acc}(s) \ge th_{\text{acc}} \}
    \label{eq:optim_lat}
\end{equation}
or optimise for $accuracy$, given a $cost$ constraint $th_{\text{cost}}$:
\begin{equation}
    \small
    s^{\star} = \arg\max_{s \in \mathcal{S}} \ \{ \text{acc}(s) \ | \ \text{cost}(s) \le th_{\text{cost}} \} 
    \label{eq:optim_acc}
\end{equation}
%

%Most importantly, our two-stage training scheme of Sec.~\ref{sec:pre-training} allows all trained exits to be interchangeably attached to the same backbone for inference. This allows for an extremely efficient search of an overly complex space, avoiding the excessive search times of NAS approaches~\cite{chen2018searching}. 

Importantly, a combination of design choices render the \textit{exhaustive exploration} of the search space not only computationally tractable, but extremely efficient. Conversely to heuristic alternatives, this guarantees \textit{optimality} within the examined space. The main enabling factors include: \textit{i)}~the informed outlining of the search space (being compact and tailor-made for segmentation), \textit{ii)}~the proposed two-stage training scheme (allowing all exits architectures to exploit a shared backbone), \textit{iii)}~a vast pruning of configurations at search time (prioritising the less costly constraint verification on latency before evaluating accuracy), and \textit{iv)}~prediction memoisation (eliminating duplicate inference execution by storing and combining per-exit predictions on the calibration set). Finally, in contrast to NAS methods~\cite{chen2018searching}, MESS overprovisioned networks are fully trained and can be customised without the need of fine-tuning, offering rapid post-training adaptation.

\subsection{Input-Dependent Exit Policy}
\label{sec:policy}

%In certain cases, the computational budget is not externally defined by a hard limit (\textit{i.e.} \textit{budgeted/ anytime inference}), but we desire to minimise the amount of computation so as to correctly classify an input sample based on its intricacy (\textit{i.e.~input-dependent inference}).
% \steve{In certain cases, the computational budget is not externally defined by a hard limit (\textit{i.e.} \textit{budgeted/ anytime inference}), but we desire to minimise the amount of computation so as to correctly classify an input sample based on its intricacy (\textit{i.e.~input-dependent inference}).} \stelios{Done.}

% In some cases, the desired optimisation objectives may be best met by combining different exit responses in an input-dependent manner.
%\noindent\textbf{Early-exit Criterion.}
During input-dependent inference, each input image goes through the selected early exits of the deployed MESS instance sequentially. After a prediction is produced from an exit, a mechanism to calculate its confidence is used to determine whether inference should continue to the next exit or not. In~\cite{li2017not}, \textit{each pixel} in an image is treated as an independent classification task, exiting early if its prediction confidence in an exit is high, thus yielding irregular computation paths. In contrast, our approach treats the segmentation of \textit{each image} as a single task, aiming to drive each sample through a uniform computation route. To this end, we fill a gap in literature by introducing a novel mechanism to quantify the overall confidence in semantic segmentation predictions. 

%Driven by the fact that not all inputs pose the same prediction difficulty, adaptive inference has been widely studied in image classification (Sec.~\ref{sec:relwork}). In this setting, each input sample goes through the selected early exits sequentially. After a prediction is produced from an exit, a mechanism to calculate an image-level confidence (as a metric of predicting difficulty) is used to determine whether inference should continue to the next exit or not.

%This technique remains highly unexplored in dense prediction problems, such as semantic segmentation. In~\cite{li2017not} \textit{each pixel} in an image is treated as an independent classification task, exiting early if its prediction confidence in an exit is high, thus yielding irregular computation paths. In contrast, our approach treats the segmentation of \textit{each image} as a single task, aiming to drive each sample through a uniform computation route. To this end, we fill a gap in literature by introducing a novel mechanism to quantify the overall confidence in semantic segmentation predictions. 

%\noindent\textbf{Confidence-tuning for MESS Networks.}~

\vspace{1mm}\noindent\textbf{Confidence Metric.}
Given the per-pixel confidence map, calculated from the probability distribution across classes of each pixel \mbox{$\boldsymbol{c}^{\text{map}} = f_c(\boldsymbol{y}) \in [0,1]^{R \times C}$} (where $f_c$ is usually $top1(\cdot)$ \cite{msdnet2018iclr} or $entropy(\cdot)$~\cite{cheng2019leveraging}), 
% or $\text{bvsb}(\cdot)$
we introduce a mechanism to reduce these \textit{every-pixel} confidence values to a single \textit{per-image} confidence. The proposed metric 
considers the \textit{percentage of pixels with high prediction confidence} (i.e. surpassing a tunable threshold $th_{i}^{\text{pix}}$) in the output of an exit $\boldsymbol{y}_i$:
\begin{equation}
\label{eq:confidence}
\small
%\sum\nolimits
    c^{\text{img}}_{i} = \frac{1}{RC}\sum_{r=1}^{R}\sum_{c=1}^{C} \mathds{1}( \boldsymbol{c}^{\text{map}}_{r,c}(\boldsymbol{y}_i) \ge th_{i}^{\text{pix}})
\end{equation}
%avoiding the naive approach of calculating the mean confidence, which is more sensitive to outliers. 

\noindent\textbf{Edge Confidence Enhancement.}
Moreover, it has been observed that due to the progressive downsampling of the feature volume in CNNs, some spatial information is unavoidably lost. As a result, semantic predictions near object edges are naturally under-confident~\cite{vu2019advent}. Driven by this observation, we enhance our proposed metric to account for these expected low-confidence pixel-predictions, by introducing a pre-processing step for $c^{\text{img}}_{i}$. Initially, we conduct edge detection on the semantic masks, followed by an erosion filter with kernel equal to %the feature volume spatial downsampling rate (8$\times$), 
the output stride of the respective exit $os_i$, in order to compute a semantic-edge map (Eq.\ref{eq:edge_det}). % $\mathcal{M} = \text{erode}( \text{cannyEdge}( \boldsymbol{\hat{y}}_i ) ,s_i)$.
Thereafter, we apply a median-based smoothing on the confidence values of pixels lying on the semantic edges (Eq.\ref{eq:final_conf_map}).
\begin{equation}
 \mathcal{M} = \text{erode}( \text{cannyEdge}( \boldsymbol{\hat{y}}_i ) ,s_i)
 \label{eq:edge_det}
\end{equation}
\vspace{-3mm}
\begin{equation}
 \widehat{\boldsymbol{c}^{\text{map}}_{r,c}}(y_i) = 
     \begin{cases}
      %median(c^{map}_{i-w:i+w,j-w:j+w}), & \text{if}\ (i,j) \in \mathcal{M}  \\
      \text{median}(\boldsymbol{c}^{\text{map}}_{w_r,w_c}(\boldsymbol{y}_i)) & \text{if}\ \mathcal{M}_{r,c}=1  \\
      \boldsymbol{c}^{\text{map}}_{r,c}(\boldsymbol{y}_i) & \text{otherwise}
    \end{cases}
    \label{eq:final_conf_map}
\end{equation}
%where $w_l$=\{$l$-$2$\cdot$os_i,...,l$+$2$\cdot$os_i$\} 
where $ w_l= \{l-2 \cdot os_i,...,l+2 \cdot os_i \} $ 
is the window size of the filter. \textit{This sets the pixels around semantic edges to inherit the confidence of their neighbouring pixel predictions}.

\vspace{1mm}
\noindent\textbf{Exit Policy.}
At inference time, each sample is sequentially processed by the selected early exits. For each prediction $\boldsymbol{y}_i$, the proposed metric $c_i^{\text{img}}$ is calculated, and a tunable confidence threshold (exposed to the search space) determines whether the sample will exit early ($c^{\text{img}}_{i} \ge th^{\text{img}}_{i}$) or be processed further by subsequent backbone layers/exits. 

%%%%%%%%%%%%%%%%%%%%%%%
\section{Evaluation}
\label{sec:experiments}
In this section, we evaluate various aspects of MESS and discuss the key benefits compared to baselines and state-of-the-art approaches. First, we present the comparative advantage of MESS over single-exit, multi-exit and NAS segmentation approaches from the literature (Sec. \ref{sec:eval_sota_seg}). We then dive deeper into the training of MESS networks (Sec. \ref{sec:eval_training}). Next, we present customised outputs for each inference setting, showcasing the effectiveness our ``train-once-deploy-everywhere'' approach (Sec. \ref{sec:eval_deployment}). Finally, we provide a qualitative analysis of some properties of MESS networks (Sec. \ref{sec:eval_qualitative}).

\vspace{-1mm}
\subsection{Experimental Setup\footnote{More extensively discussed in the Appendix.}} 

\noindent
\textbf{Models \& Datasets.}
We apply our methodology on top of DRN-50~\cite{yu2017dilated}, DeepLabV3~\cite{chen2017rethinking} and SegMBNetV2~\cite{sandler2018mobilenetv2} segmentation CNNs, using ImageNet~\cite{deng2009imagenet} pre-trained ResNet50~\cite{he2016deep} and MobileNetV2~\cite{sandler2018mobilenetv2} backbones, representing  \textit{high-end} and \textit{edge} use-cases, respectively.
% 
% \noindent
% \textbf{Datasets \& Training Configuration.} 
We train all backbones on MS COCO~\cite{lin2014microsoft} and fine-tune early exits on MS COCO and PASCAL VOC~                          \cite{everingham2010pascal} (augmented from \cite{hariharan2011semantic}) independently.

\vspace{1mm}\noindent
\textbf{Development \& Deployment Setup.} 
MESS networks are implemented on \textit{PyTorch~(v1.6.0)}. %, on top of \textit{torchvision (v0.6.0)}. % reference segmentation networks. 
% A total of 40 Nvidia GTX 2080Ti GPUs were used for training multiple instances of the proposed search space.
% 
% \noindent
% \textbf{Deployment.}
For inference, we deploy MESS instances on a \textit{high-end} (Nvidia GTX1080Ti; 250W TDP) and an \textit{edge} (Nvidia Jetson AGX Xavier; 30W TDP) compute platform. 

% \steve{Define baselines early-on.} \stelios{Done.}

\vspace{1mm}\noindent
\textbf{Baselines.} %$^{\ref{note1}}$
To compare our work against the following state-of-the-art baselines:

\noindent
\textit{1)}~\textbf{DRN}~\cite{yu2017dilated}; 
\textit{2)}~\textbf{DLBV3}~\cite{yu2017dilated,chen2018encoder}; 
\textit{3)}~\textbf{segMBNetV2}~\cite{sandler2018mobilenetv2}; 
\textit{4)}~\textbf{LC}~\cite{li2017not}; 
\textit{5)}~\textbf{AutoDLB}~\cite{liu2019auto};
\textit{6)}~\textbf{E2E}~\cite{msdnet2018iclr,teerapittayanon2016branchynet}; 
\textit{7)}~\textbf{Frozen}~\cite{sdn2019icml,hapi2020iccad};
\textit{8)}~\textbf{KD}~\cite{hinton2015distilling} and
\textit{9)}~\textbf{SelfDistill}~\cite{scan2019neurips,phuong2019distillation,zhang2019be}. \vspace{2mm}

\subsection{MESS End-to-End Evaluation}
\label{sec:eval_sota_seg}
\noindent\textbf{Comparison with Single-Exit Baselines.}
First, we apply our MESS framework on single-exit segmentation backbones from the literature, namely \textbf{DRN}, \textbf{DLBV3} and \textbf{segMBNetV2}. Table~\ref{tab:mess} lists the achieved results for MESS instances optimised for varying use-cases (framed as speed/accuracy constraints fed to our configuration search).

For a DRN-50 backbone on MS COCO, we observe that a latency-optimised MESS instance achieves a 3.36$\times$ workload reduction with no accuracy drop (row~(iii)), translating to a latency speedup of 2.23$\times$ over the single-exit \textbf{DRN} (row~(i)). This improvement is amplified to 4.01$\times$ in workload (2.65$\times$ in latency) for use-cases that can tolerate a controlled accuracy degradation of $\le$1 pp (row~(iv)). Additionally, a MESS instance optimised for accuracy under the same workload budget as \textbf{DRN}, can achieve an mIoU gain of 5.33~pp compared to \textbf{DRN}, with 1.22$\times$ fewer GFLOPs (row~(ii)).%, restricted to use only \textit{FCN-head}.

Similar results are obtained for \textbf{DLBV3}, as well as when targeting PASCAL VOC. Moreover, the gains are consistent on \textbf{segMBNetV2}, %(rows (ix)-(xii)), 
which forms an inherently efficient segmentation model, with 15.7$\times$ smaller workload than DRN-50. This demonstrates the model-agnostic nature of our framework, yielding complementary gains to efficient backbone design, by exploiting the orthogonal dimension of input-dependent inference.

\vspace{1.5mm}\noindent\textbf{Comparison with Multi-Exit Baselines.}
Next, we compare MESS networks against \textit{Deep Layer Cascade} (\textbf{LC})~\cite{li2017not}, the current SOTA in multi-exit segmentation, which proposes \textit{per-pixel} early-exiting through multiple segmentation heads. Due to their unstructured computation, standard BLAS libraries cannot realise true latency benefits from this approach. However, we apply \textbf{LC}'s pixel-level exit policy on diverse MESS configurations, and compare with our image-level policy analytically (in GFLOPs), tuning both thresholds so as to meet varying accuracy requirements.

\begin{figure*}[t]
%\vspace{-3mm}
      \centering
      \captionof{table}{\scriptsize End-to-end evaluation of MESS network designs \alex{DeepLabV3 on PASCAL VOC results to be updated following new training protocol.}}
      \vspace{-3mm}
      \renewcommand{\arraystretch}{1.0} 
      \setlength\tabcolsep{3pt} % default value: 6pt
        { 
        \resizebox{0.84\columnwidth}{!}{
        \begin{tabular}{lccccccrccrc}
         \hline 
            \multirow{2}{*}{\textbf{Method}} & & \multirow{2}{*}{\textbf{Backbone$^*$}} & 
            \multirow{2}{*}{\textbf{Head}}  & \multicolumn{2}{c}{ \textbf{Search Targets} } &  \multicolumn{3}{c}{ \textbf{Results: MS COCO } } & \multicolumn{3}{c}{ \textbf{Results: PASCAL VOC} }  \\
            %\hline
             & & & & Error & GFLOPs & mIoU & GFLOPs & Latency$^{\dag}$ & mIoU & GFLOPs & Latency$^{\dag}$  \\
             \hline
             \textbf{DRN}~\cite{yu2017dilated} & (i) &
              ResNet50 & FCN & \multicolumn{2}{c}{ --Baseline--} & 59.02\% & 138.63 & 39.96ms & 72.23\% & 138.63 & 39.93ms  \\
             \textbf{Ours} & (ii) & 
             ResNet50 & FCN & min & $\le$ 1$\times$ & 64.35\% & 113.65 & 37.53ms  & 79.09\% & 113.65 & 37.59ms  \\
             \textbf{Ours} & (iii) &
             ResNet50 &  FCN & $\le$ 0.1\% & min  & 58.91\% & 41.17 & 17.92ms  & 72.16\% & 44.81 & 18.63ms  \\
             \textbf{Ours} & (iv) &
             ResNet50 &  FCN & $\le$ 1\% & min  & 58.12\% & 34.53 & 15.11ms  & 71.29\% & 38.51 & 16.80ms  \\
             \hline
             \textbf{DLBV3}~\cite{chen2017rethinking} & (v) &
              ResNet50 &  DLB &\multicolumn{2}{c}{ --Baseline--} & 64.94\% & 163.86 & 59.05ms & 80.32\% & 163.86 & 59.06ms  \\
             \textbf{Ours} & (vi) &
              ResNet50 & DLB & min & $\le$ 1$\times$ & 65.52\% & 124.10 & 43.29ms  & 82.32\% & 124.11 & 43.30ms  \\
             \textbf{Ours} & (vii) &
             ResNet50 & DLB &  $\le$ 0.1\% & min & 64.86\% & 69.84  & 24.81ms  & 80.21\% & 65.29 & 24.14ms  \\
             \textbf{Ours} & (viii) &
             ResNet50 & DLB &  $\le$ 1\% & min & 64.03\% & 57.01 & 20.83ms & 79.30\% & 50.29 & 20.11ms  \\
             \hline
             \textbf{segMBNetV2}~\cite{sandler2018mobilenetv2} & (ix) &
              MobileNetV2 &  FCN & \multicolumn{2}{c}{ --Baseline--} & 54.24\% & 8.78 & 67.04ms & 69.68\% & 8.78 & 67.06ms   \\
             \textbf{Ours} & (x) &
              MobileNetV2 & FCN & min & $\le$ 1$\times$ & 57.49\% & 8.10 &  56.05ms  & 74.22\% & 8.10 & 56.09ms \\
             \textbf{Ours} & (xi) &
              MobileNetV2 & FCN & $\le$ 0.1\% & min  & 54.18\% & 4.05  & 40.97ms & 69.61\% & 3.92  & 32.79ms  \\
             \textbf{Ours} & (xii) &
             MobileNetV2 & FCN &  $\le$ 1\% & min  & 53.24\% & 3.48 & 38.83ms & 68.80\% & 3.60 & 31.40ms  \\
             \hline
             \multicolumn{12}{c}{$^*$Dilated network \cite{yu2017dilated} based on backbone CNN. %} %\\
              %\multicolumn{7}{c}{
              $^{\dag}$Measured on: GTX for ResNet50 and AGX for MobileNetV2 backbone.}
              \vspace{-3mm}
             \label{tab:mess} 
        \end{tabular}
        }}
\vspace{-2mm}
\end{figure*}

\begin{figure}[t]
    \begin{minipage}[t]{0.43\textwidth}
        \centering
        \captionof{table}{\scriptsize Comparison with LC (speedup to backbone)}
        \vspace{-3mm}
        \renewcommand{\arraystretch}{1.0} 
        \setlength\tabcolsep{3pt} % default value: 6pt
        { 
        \resizebox{1\columnwidth}{!}{
        \begin{tabular}{lllcc}
          \hline
           \multirow{2}{*}{\textbf{Head}} & \multirow{2}{*}{\textbf{Search Target}} &  \multicolumn{1}{c}{\multirow{2}{*}{\textbf{Exit Points}}} & \multicolumn{2}{c}{ \textbf{Exit Policy} } \vspace{-0.2mm} \\
             & & &  \small{LC\cite{li2017not}} & \small{Ours} \\
            \hline
            FCN & Error $\le$ \phantom{0}0.1\% & \textit{3-exit:}\{$\mathcal{E}_1,\mathcal{E}_3,\mathcal{E}_6$\} & 1.13$\times$ & \textbf{3.36$\times$} \\
            %FCN & Error $\le$ 1.0\% & \textit{3-exit:}\{$\mathcal{E}_1,\mathcal{E}_2,\mathcal{E}_6$\} & X.XX$\times$ & \textbf{4.01$\times$} \\
            FCN & Error $\le$ 10.0\%  & \textit{2-exit:}\{$\mathcal{E}_1,\mathcal{E}_6$\} & 0.98$\times$ & \textbf{6.02$\times$} \\ 
           \hline
           \label{tab:lc}
        \end{tabular}
        }
        }
        %\vspace{-7.5mm}   
    \end{minipage}
    \hfill  
    \begin{minipage}[t]{0.54\textwidth}
      \centering
      \captionof{table}{\scriptsize Comparison with SOTA NAS approach}
      \vspace{-3mm}
      \renewcommand{\arraystretch}{1.0} 
      \setlength\tabcolsep{3pt} % default value: 6pt
        { 
        \resizebox{1\columnwidth}{!}{
        \begin{tabular}{lccrcccr}
          \hline
           \multirow{2}{*}{\textbf{Method}} & \multirow{2}{*}{\textbf{Approach}} & \multirow{2}{*}{\textbf{ImgNet}} & \multirow{2}{*}{\textbf{Training$^*$}} & \multicolumn{2}{c}{\textbf{Adaptation$^*$} } & \multirow{2}{*}{ \textbf{mIoU}} & \multirow{2}{*}{\textbf{GFLOPs}}  \vspace{-1mm}  \\
           & & &  & \scriptsize{search} & \scriptsize{re-training} & &\\   
           \hline
           \textbf{DLBV3}~\cite{chen2017rethinking} & Baseline & \checkmark   & 192 & \multicolumn{2}{c}{ \small -Non-adaptive-} & 80.32\% & 163.86 \\   
           \textbf{AutoDLB}~\cite{liu2019auto} & NAS &  - & 12,248 & 72 & 12,176 & 79.78\% 
           & \phantom{1}57.61 \\
           \textbf{Ours} & MESS &  \checkmark  & 2,580 & $<$1 & - & 79.94\% & \phantom{1}51.59  \\    %updated after fine-tuning 
           %\textbf{Ours} & MESS &  \checkmark  & 2,580 & $<$1 & - & 79.40\% & \phantom{1}74.20  \\  % old
           \hline
           \multicolumn{8}{c}{ $^*$Initial-training and Adaptation times expressed in GPU-hours.}
           \label{tab:nas}
        \end{tabular}
        }
        }
        %\vspace{-7mm}
    \end{minipage}
\end{figure}

By using SOTA techniques for semantic segmentation, such as larger dilation rates or DeepLab's ASPP, the gains of \textbf{LC} rapidly fade away, as for each pixel that propagates deeper on, a substantial feature volume %falls within its receptive field and 
needs to be precomputed. Concretely, when employing \textbf{LC} on our designs, up to a substantial 45\% of the feature volume at the output of the first exit falls within the receptive field of a \textit{single pixel} in the final output for the case of \textit{FCN-Head}, reaching 100\% for \textit{DLB-Head}. As a result, %when employing \textbf{LC}'s exit-policy in MESS instances, 
\textbf{LC}'s policy presents heavily dissipated to no reduction in workload against the corresponding single-exit baselines, being heavily reliant on the exit placement. In contrast, the respective MESS instances equipped with our proposed exit policy (Sec.~\ref{sec:policy}) are able to achieve significant workload reduction, reported in Table~\ref{tab:lc}.

\vspace{1.5mm}\noindent\textbf{Comparison with NAS Baselines.}
Finally, we position our work against NAS solutions for deriving efficient segmentation models. We employ Auto-DeepLab (\textbf{AutoDLB})~\cite{liu2019auto} as our strong baseline, due to its SOTA performance both in accuracy and search efficiency, and use our framework to generate a MESS instance matching its accuracy (staring from  DeepLabV3~\cite{chen2017rethinking} backbone). Table~\ref{tab:nas} lists our findings on PASCAL VOC. % using a $<$1 pp accuracy drop constraint compared to the original DeepLabV3~\cite{chen2017rethinking}, for both methods.

%Both \textbf{AutoDLB} and MESS frameworks could meet the given accuracy constraint.
Remarkably, MESS achieves a better (but comparable) speed-accuracy trade-off than \textbf{AutoDLB} (3.17$\times$ vs 2.85$\times$ speedup over DeepLabV3), although the latter samples from a larger space during search ($10^{19}$ points vs $10^6$) and takes advantage of more degrees of freedom during training. Additionally, being able to exploit ImageNet pre-trained backbones, MESS demonstrates significant training time savings (4.7$\times$ faster) compared to NAS-crafted models like \textbf{AutoDLB}, that can only be trained from scratch. 
% pays a much higher training cost, due to the fact that NAS-crafted models need to be trained from scratch. In contrast, MESS can , resulting to significant training time savings . 

%Additionally, \textbf{AutoDLB} pays a much higher training cost, due to the fact that NAS-crafted models need to be trained from scratch. In contrast, MESS can exploit ImageNet pre-trained backbone networks, resulting to significant training time savings (4.7$\times$ faster). 

Most importantly, due to our ``train-once-deploy-everywhere'' design, enabled by the two-stage training approach of MESS, %(which decouples the training of early exits from the backbone), 
 after the initial training of the overprovisioned MESS network \textit{all $\mathit{10^6}$ possible MESS instances are ready-to-deploy} without any need for re-training. Alternatively, training end-to-end all $10^6$ MESS instances would require \mbox{$>$200 million} GPU-hours. As a result, different MESS instances can be obtained with a minimal search cost ($<$1 GPU-hour). Overall, MESS offers up to five orders of magnitude faster adaptation time compared to NAS-based methodologies. %, which need to run a complete training from scratch to validate every candidate instance.

\subsection{MESS Training Evaluation}
\label{sec:eval_training}
Having shown the benefits of MESS networks against different state-of-the-art methods, we now move to the evaluation of specific components of our framework. %,comparing them with alternative techniques available in the literature.

% \subsection{MESS Training Evaluation}
%First, we assess the performance of our proposed scheme for training MESS networks and provide an in-depth evaluation of our pre-training (Section~\ref{sec:eval_exit_aware_pretraining}) and distillation (Section~\ref{sec:eval_positive_filtering_distillation}) techniques in terms of accuracy.% and performance.

\begin{figure}[t]
   %\vspace{-3mm}
    \begin{minipage}[t]{0.65\textwidth}
      \centering
      \captionof{table}{\scriptsize Evaluation of two-stage training scheme on DRN-50 (mIoU).}
      \vspace{-3mm}
      \renewcommand{\arraystretch}{1.0} 
      \setlength\tabcolsep{3pt} % default value: 6pt
        { 
        \resizebox{1\columnwidth}{!}{
        \begin{tabular}{ lcccccccc }
            \hline
            \textbf{Method} & \textbf{Init.} & \textbf{Loss} & $\mathcal{E}_1$ & $\mathcal{E}_2$ & $\mathcal{E}_3$ & $\mathcal{E}_4$ & $\mathcal{E}_5$ & $\mathcal{E}_6$ \\
            %\multicolumn{6}{c}{ \textbf{DRN-50} } \\
            %\hline
            %&
            %\multicolumn{2}{c|}{} & $\mathcal{E}_1$ & $\mathcal{E}_2$ & $\mathcal{E}_3$ & $\mathcal{E}_4$ & $\mathcal{E}_5$ & $\mathcal{E}_6$ \\
            \hline
           (i)\phantom{ii} Baseline Init.  & ImageNet & $\mathcal{L}_{\text{CE}}(\mathcal{E}_6)$                  &   -  & - & -& -& - & 59.02\%  \\
           (ii)\phantom{i} \textbf{E2E}~\cite{msdnet2018iclr,teerapittayanon2016branchynet} & ImageNet &   $\mathcal{L}_{\text{CE}}(\mathcal{E}_1)$+...+$\mathcal{L}_{\text{CE}}(\mathcal{E}_6)$        &   29.02\% & 40.67\% & 48.64\% & 51.69\% & 55.34\% & 58.33\%  \\
           (iii) Exit-aware Init. & ImageNet & Eq.~(\ref{eq:drop}) (\textbf{Ours}) &   28.21\% & 39.61\% & 47.13\% & 50.81\% & 56.11\% & \textbf{59.90\%}  \\
            (iv) \textbf{Frozen}~\cite{sdn2019icml,hapi2020iccad} & (i) & $\mathcal{L}_{\text{CE}}(\mathcal{E}_1)$, {\scriptsize ...}, $\mathcal{L}_{\text{CE}}(\mathcal{E}_5)$      &   23.94\%  & 31.50\% & 38.24\% & 44.73\% & 54.32\% & 59.02\%  \\  
            (v)\phantom{i} \textbf{Ours ({\small \S\ref{sec:pre-training}})} & (iii) & $\mathcal{L}_{\text{CE}}(\mathcal{E}_1)$, {\scriptsize ...}, $\mathcal{L}_{\text{CE}}(\mathcal{E}_5)$      &   \textbf{32.40\% } & \textbf{43.34\%} & \textbf{50.81\%} &\textbf{ 53.73\%} & \textbf{57.9\%} & \textbf{59.90\%}  \\ 
            \hline
            \multicolumn{9}{c}{$^*$ Experiments repeated 3 times. The sample stdev in mean IoU is at most  $\pm$ 0.09 in all cases. }
            \label{tab:pre-train} 
        \end{tabular}
        }
        }
        \vspace{-3mm}
        \centering
        \captionof{table}{\scriptsize Evaluation of Positive Filtering Distillation (mIoU)}
        \vspace{-0mm}
        \renewcommand{\arraystretch}{1.0} 
        \setlength\tabcolsep{3pt} % default value: 6pt
        { 
        \resizebox{0.8\columnwidth}{!}{
        \begin{tabular}{ ll|ccc|ccc}
        \hline
        \multirow{2}{*}{\textbf{Method}} &
        \multirow{2}{*}{\textbf{Loss}} & \multicolumn{3}{c|}{ \textbf{DRN-50} } & \multicolumn{3}{c}{ \textbf{MobileNetV2} } \\
        %\hline
        & 
        & $\mathcal{E}_1$ & $\mathcal{E}_2$ & $\mathcal{E}_3$ & $\mathcal{E}_1$ & $\mathcal{E}_2$ & $\mathcal{E}_3$ \\
        \hline
        \textbf{E2E}~\cite{msdnet2018iclr,teerapittayanon2016branchynet} & 
        CE & 49.96\% & 55.40\% & 58.96\% &    31.56\% & 41.57\% & 51.59\%   \\
        \textbf{KD}~\cite{hinton2015distilling} & 
        KD &  50.33\% & 55.67\% & 59.08\% &    31.04\% & 41.93\% & 51.66\%   \\
        \textbf{SelfDistill}~\cite{scan2019neurips,phuong2019distillation,zhang2019be} &
        CE+KD & 50.66\% & 55.91\% & 58.84\% &    32.08\% & 41.96\% & 51.58\%   \\
        \textbf{Ours ({\small \S\ref{sec:distillation}})} & 
        PFD  &  \textbf{51.02\%} & \textbf{56.21\%} & \textbf{59.36\%} &   \textbf{ 33.36\%} & \textbf{42.95\%} & \textbf{52.20\%}   \\
        \hline
        \multicolumn{8}{c}{ CE=Cross-entropy, KD=Knowledge Distillation, PFD=Positive Filtering Distillation}
        \label{tab:distilation} 
        \end{tabular}
        }
        }
    %\vspace{-7.5mm}   
    \end{minipage}
    \hfill  
  \begin{minipage}[t]{0.33\textwidth}
      \centering
      \vspace{4mm}
      \includegraphics[align=t, trim =0mm 8mm 0mm 16mm, clip,width=0.98\columnwidth]{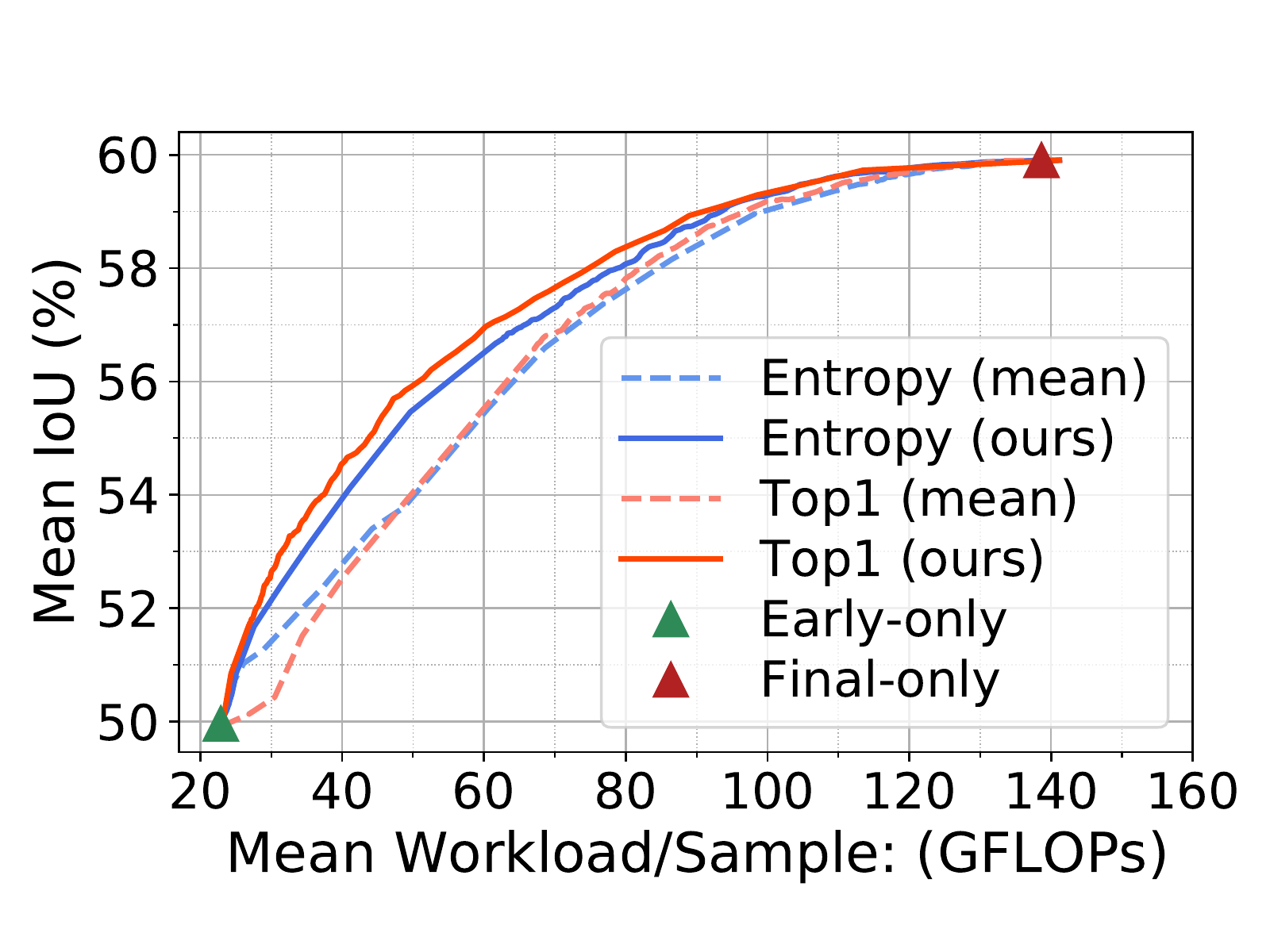}
      \vspace{-3mm}
      \caption{\scriptsize  Comparison of different early-exit policies on a DRN-50-based MESS instance with two exits.}
      %\vspace{-5mm}
      \label{fig:confmetric}    
  \end{minipage}
  \vspace{-3mm}
\end{figure}

\vspace{1mm}\noindent\textbf{Exit-Aware Pre-training.}
%\label{sec:eval_exit_aware_pretraining}
Initially, we demonstrate the effectiveness of the proposed training scheme. We compare the accuracy of models with uniform exit configuration across all candidate exits points, trained using different strategies. %including state-of-the-art end-to-end (\textbf{E2E SOTA}) and frozen-backbone (\textbf{Frozen SOTA}) methodologies.
Table~\ref{tab:pre-train} summarises the results of this comparison on a DRN-50 backbone with $N$=$6$, on MS COCO. %validation set. %Rows (i)-(iii) constitutes end-to-end training approaches, whereas rows (iv)-(v) assumes a pre-trained backbone with its weights ``frozen". %, where multiple exits can be trained independently. 
%The top cluster (rows (i)-(iii)) constitutes end-to-end training approaches, whereas the bottom cluster (rows (iv)-(v)) assumes a pre-trained backbone with its weights ``frozen", where multiple exits can be trained independently.

When multiple exits are attached to the backbone and jointly trained end-to-end, as in \cite{msdnet2018iclr,teerapittayanon2016branchynet}, the accuracy of the final exit can notably degrade (row (ii)) compared to a vanilla training of the backbone with solely the final exit attached (row(i)). This is attributed to contradicting gradient signals between the early and the late classifiers and to the larger losses of the early results, which dominate the loss function~\cite{adaptive2017icml}. On the other hand, freezing the weights of the vanilla backbone of row (i) and independently training the same early exits, as in \cite{sdn2019icml,hapi2020iccad}, leads to degraded accuracy in shallow exits (row (iv)). This is due to the limited degrees of freedom of this second training stage and the weaker semantics extracted by shallow layers of the frozen backbone. 

Our (1st-stage) exit-aware pre-training %comprises an \textit{exit-dropout loss} that only trains the early exits one-by-one in an alternating fashion, along with the final exit. This approach 
pushes the extraction of semantically strong features towards shallow parts of the network, while yielding the highest accuracy on the final exit (row (iii)). Similar to observations from~\cite{43022}, we fathom that the extra signal midway through the model acts both as a regulariser and as an extra backpropagation source, reducing the effect of vanishing gradients. 

Capitalising on this exit-aware pre-trained backbone, and without any harm of the final exit's accuracy, our subsequent frozen backbone training achieves consistently higher accuracy (up to 12.57pp) across all exits (row (v)) compared to a traditionally pre-trained segmentation network (\textbf{Frozen}), and up to 3.38~pp compared to an end-to-end trained model (\textbf{E2E}), which also suffers a \mbox{1.57 pp} accuracy drop in the final exit.

\vspace{1mm}\noindent\textbf{Positive Filtering Distillation.} 
%\label{sec:eval_positive_filtering_distillation}
Here, we quantify the benefits of Positive Filtering Distillation (PFD) for the second stage (frozen-backbone) of our training methodology. 
% In this section, we conduct further experiments to quantify the benefits of our Positive Filtering Distillation (PFD) scheme for the second stage (frozen-backbone) of our training methodology. 
To this end, we compare against \textbf{E2E} utilising cross-entropy loss (CE), traditional knowledge distillation (\textbf{KD}), and \textbf{SelfDistill} approach commonly used in multi-exit classification.
%employing a combined loss (CE+KD).
% To this end, we compare against the baseline cross-entropy (CE) \stelios{This is used in MSDNet and BranchyNet.}, the traditional knowledge distillation (KD) approach \cite{hinton2015distilling}, and their combined loss (CE+KD) as used in \cite{phuong2019distillation,zhang2019be}. \stelios{CE+KD is used in the SCAN paper at NeurIPS 2019 (same authors as [69]), in case we want to explicitly say that we compare with them.} 
% To this end, we compare against the baseline cross-entropy (CE) and the traditional knowledge distillation (KD) approach \cite{hinton2015distilling}, as it is employed in early-exit classification networks \cite{phuong2019distillation,zhang2019be} (CE+KD). 
Table \ref{tab:distilation} summarises our results on a representative exit-architecture, on both DRN-50 and MobileNetV2, across MS COCO validation set.

% When employing our proposed loss, we observe a notable improvement \steve{Just put here a number for how better PFD is and we can keep this section concise.} in accuracy across all studied cases. 
Our proposed loss consistently yields higher accuracy across all cases, achieving up to 1.8, 2.32 and 1.28 pp accuracy gains over \textbf{E2E}, \textbf{KD} and \textbf{SelfDistill}, respectively. % CE, KD and CE+KD, respectively. 
This accuracy boost is more salient on shallow exits, %concentrating the training process on ``easy" pixels, 
whereas a narrower improvement is obtained in deeper exits where the accuracy gap to the final exit is natively bridged. 

\vspace{-1.5mm}
\subsection{MESS Deployment under Different Settings}
\begin{figure}[t]
   %\vspace{-4mm}
   \begin{minipage}[t]{0.53\textwidth}
        \centering
        \begin{subfigure}[t]{0.532\linewidth}
            \centering
        	\includegraphics[trim =1mm 0mm 3mm 0mm, clip,width=1\linewidth]{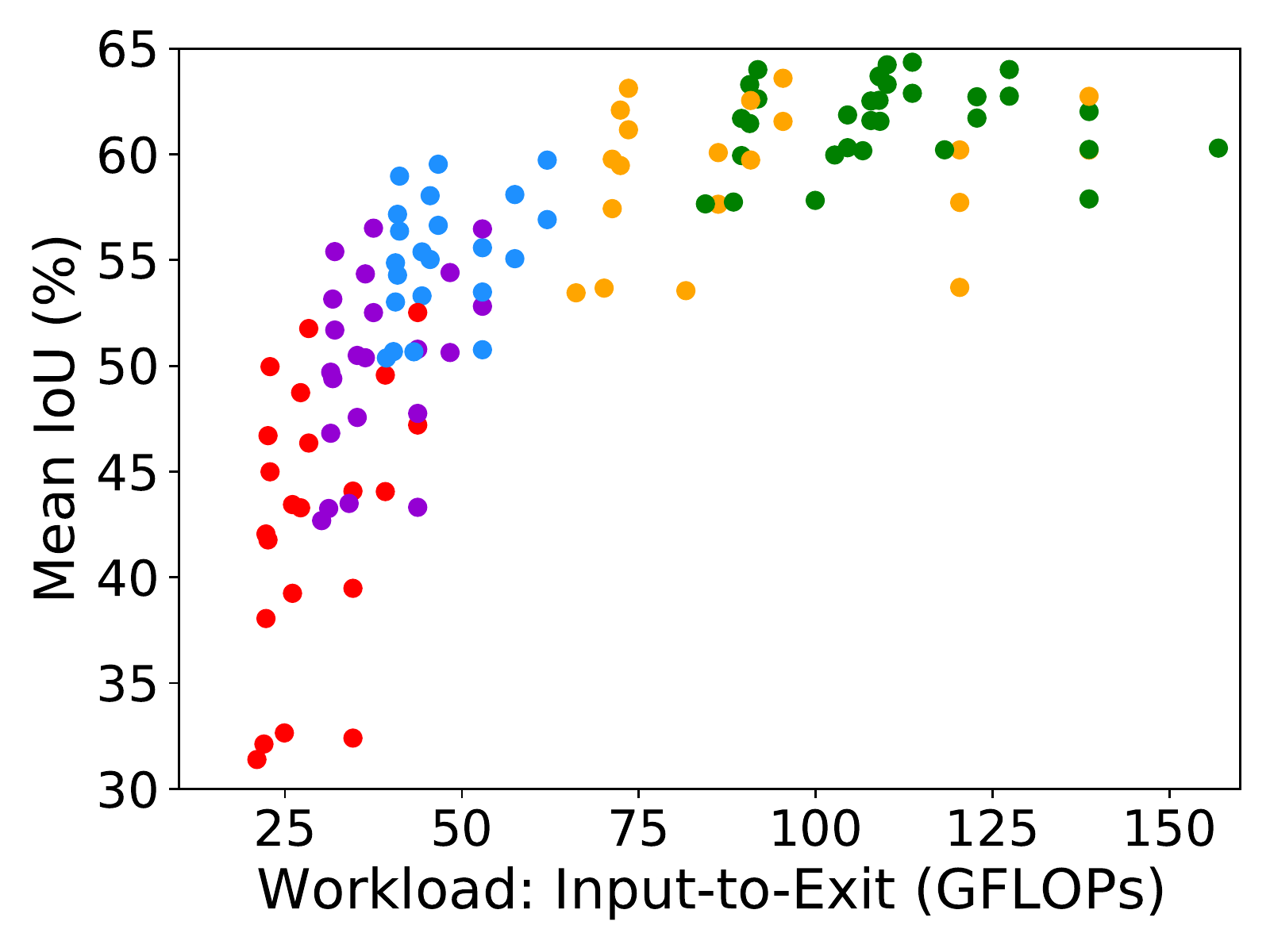}
        	\label{fig:space_input2exit}
        \end{subfigure}%
        ~ 
        \begin{subfigure}[t]{0.465\linewidth}
            \centering
        	\includegraphics[trim =22mm 0mm 2mm 2mm, clip,width=1\linewidth]{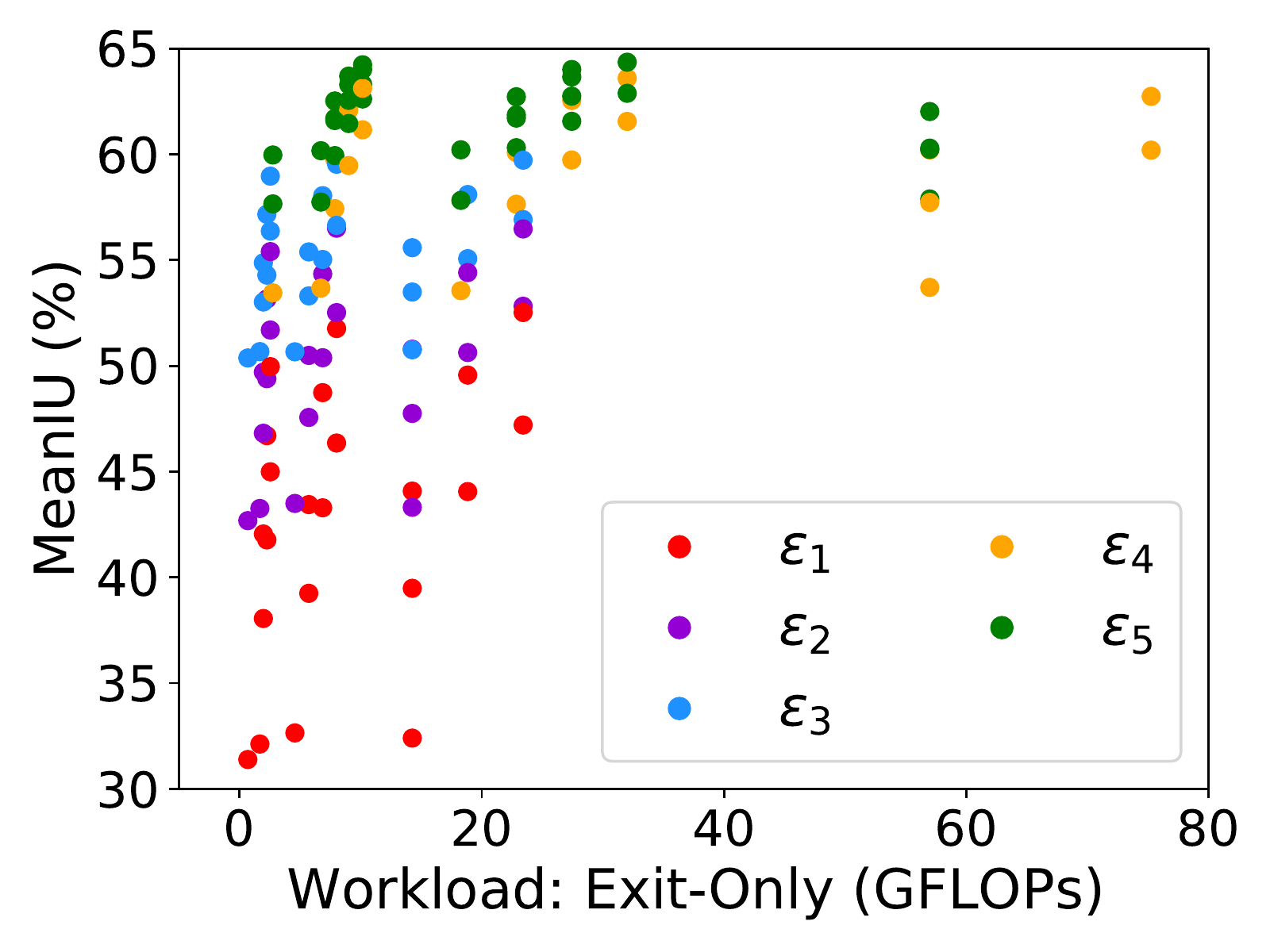}
        	\label{fig:space_exitonly}
        \end{subfigure}
        \put(-180,3){\scriptsize (a)}
    	\put(-89,3){\scriptsize (b)}
        \vspace{-7mm}
        \caption{\scriptsize Workload-accuracy trade-off between design points.  Capturing: (a) input-to-exit workload, %examined in budgeted inference, 
        (b) the overhead of each exit. % used for anytime inference.
        }
        \vspace{-6.5mm}
        \label{fig:space}
  \end{minipage}
\hfill 
    \begin{minipage}[t]{0.45\textwidth}
 \vspace{-23mm}
\captionof{table}{ \scriptsize DRN-50 with one early exit optimised for different inference schemes (Requirement of 50\% mIoU)
}
\vspace{-6mm}
\begin{center}
\renewcommand{\arraystretch}{1.0} 
\setlength\tabcolsep{3pt} % default value: 6pt
{ 
\resizebox{1.0\columnwidth}{!}{
\begin{tabular}{ ccccccc}
    \hline
    &
    \multirow{2}{*}{\textbf{Inference}} & \multicolumn{3}{c}{ \textbf{Workload (GFLOPs)} } & \multicolumn{2}{c}{ \textbf{mIoU} } \vspace{-1mm}\\
    &
    & \small{Overhead} & $\mathcal{E}_{\text{early}}$ & $\mathcal{E}_{\text{final}}$ & $\mathcal{E}_{\text{early}}$ & $\mathcal{E}_{\text{final}}$  \\
    \hline %/dashline
    (i) & Final-Only    &  -   & -     & 138.63  &   -     & 59.90\% \\
    (ii) & Budgeted     & 8.01 & 28.34 & - & 51.76\% & - \\
    (iii) & Anytime & 0.69 & 39.32 & 139.33  & 50.37\% & 59.90\% \\
    %\hdashline
    \hline
    (iv) & Input-Dep. & 2.54 & \multicolumn{2}{c}{ ( $\mathcal{E}_{\text{sel}}$: 23.02 )}   & \multicolumn{2}{c}{  ( $\mathcal{E}_{\text{sel}}$: 50.03\% ) } \\
    \hline
\end{tabular}
}
}
\end{center}
\vspace{-1.5em}
\label{tab:inference} 

\end{minipage}
    
\end{figure}

\vspace{-0.5mm}
\label{sec:eval_deployment}
In this section, we showcase the effectiveness and flexibility of the proposed \textit{train-once, deploy-everywhere} approach for semantic segmentation. %, optimised under varying workload/accuracy constraints. 
There are three inference settings in MESS networks: \textit{i)}~budgeted, \textit{ii)}~anytime and \textit{iii)}~input-dependent, for which we optimise separately \textit{post-training} (Sec.~\ref{sec:deployment}). Here, we employ our search to find the best single early-exit architecture for each case, using a 50\% mIoU requirement. The results are summarised in Table~\ref{tab:inference}. Fig.~\ref{fig:space} also depicts the underlying workload-accuracy relationship across the architectural configuration space for DRN-50 backbone. Different points represent different architectures, colour-coded by their placement in the network. %Fig.~\ref{fig:space}a showcases the cost-accuracy trade-off from the input until the respective early exit of the network, which makes it relevant for the \textit{budgeted inference}. On the other hand, Fig.~\ref{fig:space}b shows only the overhead each of the attached heads is putting, which is relevant for the \textit{anytime} and \textit{input-dependent inference} modes. 
%To guide our conclusions on the best designs across the inference modes, we complement our experiments with measurements from two-exit networks,  meeting a 50\% mean IoU requirement, presented in Table~\ref{tab:inference}.

\begin{figure}[t]
    \centering
    \vspace{4mm}
	%\includegraphics[trim =0mm 125mm 50mm 3mm, clip,width=1\columnwidth]{prog_vs_any_v4}
	%\put(-235,4){\small (a)}
	%\put(-116,4){\small (b)}
	\includegraphics[trim =0mm 125mm 75mm 2mm, clip,width=0.9\columnwidth]{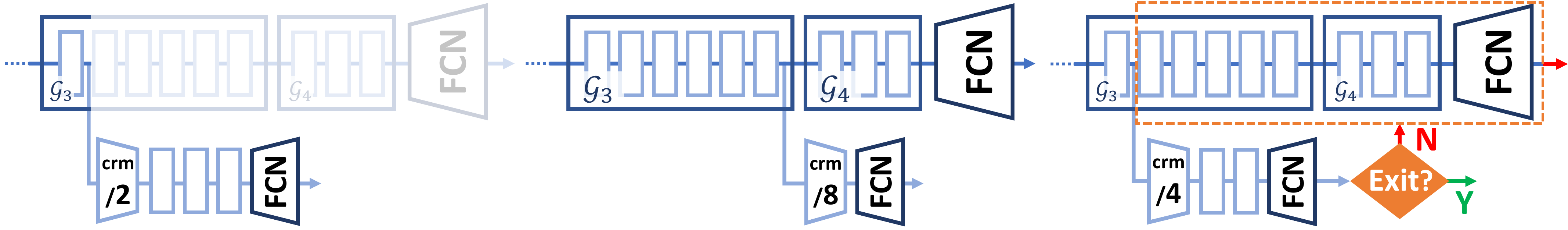}
	\put(-311,4){\scriptsize (a)}
	\put(-207,4){\scriptsize (b)}
	\put(-103.5,4){\scriptsize (c)}
	\vspace{-2mm}
	\caption{\scriptsize Selected design points for: (a) budgeted, (b) anytime and (c) input-dependent inference, with the same accuracy target.}
	%\vspace{-6mm}
	\label{fig:prog_vs_any}
\end{figure}

\vspace{1mm}\noindent\textbf{Budgeted Inference}. In this setting, we search for a \textit{single-exit submodel} that can  execute within a given latency/memory/accuracy target. Our method is able to provide the most efficient MESS instance, tailored to the requirements of the underlying application and target device (Table \ref{tab:inference}; row (ii)). This optimality gets translated in Fig.~\ref{fig:space}a, showcasing the cost-accuracy trade-off from the input until the respective early exit of the network, in the presence of candidate design points along the Pareto front of the search space. 
% In this setting, our search favours designs with the flexibility of more trainable layers mounted earlier on the network.
In this setting, our search tends to favour designs with powerful exit architectures, consisting of multiple trainable layers, mounted earlier in the network (Fig.~\ref{fig:prog_vs_any}a).
%As a result, the sub-model up to $\mathcal{E}_\text{early}$ (row 2) achieves a computational cost reduction of 4.89$\times$ over the baseline while still meeting the 50\% mean IoU target.


\noindent\textbf{Anytime Inference.} %Here, we treat a given deadline as a cut-off point of computation. When this \mbox{deadline} is met, we simply take the last output of the early exit network available -- or use this as a placeholder result to be asynchronously refined until the result is actually used. 
In this setting, each sample is sequentially processed by multiple exits, progressively refining its prediction. When a deadline is met or a result is needed, the last available output of the multi-exit network is asynchronously returned. 
%we treat a given deadline as a cut-off point of computation. When this \mbox{deadline} is met, we simply take the last output of the early exit network available -- or use this as a placeholder result to be asynchronously refined until the result is actually used. 
This paradigm creates an inherent trade-off: denser exits provide more frequent ``checkpoints", whereas each added head adds computational overhead when not explicitly used.
To control this trade-off, our method considers the additional computational cost of each exit, when populating the MESS network architecture (Fig.~\ref{fig:space}b). 
% In this setting, our search produces heads with extremely lightweight architecture and minimal computational overhead, mounted deeper in the network. 
Contrary to budgeted inference, in this setting our search produces heads with extremely lightweight architecture, sacrificing flexibility for reduced computational overhead, mounted deeper in the network (Fig.~\ref{fig:prog_vs_any}b). Table~\ref{tab:inference} showcases that, for anytime inference (row (iii)), our search yields an exit architecture with 11.6$\times$ less computational  requirements compared to budgeted inference (row (ii)), under the same accuracy constraint.

% As an example, we run our methodology under both these inference settings, introducing the same requirement of 50\% mean IU under both computational settings. Selected design points are listed in Table \ref{tab:inference}. It is evident from Fig.~\ref{fig:prog_vs_any} that under budgeted inference setting our methodology favours a design with the flexibility of more trainable layers mounted earlier on the network. In contrast, the anytime inference setting concludes to an extremely lightweight architecture with minimal computational overhead, mounted deeper in the network. 

\vspace{1mm}\noindent\textbf{Input-Dependent Inference.} In this setting, each input sample propagates through the selected MESS instance until the model  yields a confident-enough prediction ($\mathcal{E}_{\text{sel}}$). %, based on the criterion of Eq.~\ref{eq:confidence}.
%inference ends when the model is confident enough about its prediction (Eq.~\ref{eq:confidence}).
% 
% After each exit yields its prediction, the exit-policy estimates the prediction confidence via a metric. Subsequently this metric is compared with a (tunable) threshold to determine whether processing for this particular sample should continue (propagate to the next exit) or exit early (terminate processing, yielding the latest prediction as output). 
% 
By selecting different threshold values for the confidence-based exit policy (Sec. \ref{sec:policy}), even the simplest (2-exit) configuration of input-dependent MESS network (Fig.~\ref{fig:prog_vs_any}c) provides a fine-grained trade-off between workload and accuracy. Exploiting this trade-off, we observe that input-dependent inference (Table \ref{tab:inference}; row (iv)) offers the highest computational efficiency under the same (50\% mIoU) constraint. 

\textbf{\textit{Confidence Metric:}} 
To evaluate MESS exit-policy, we apply the proposed \textit{image-level} confidence metric for segmentation, on top of both \textit{top1} \cite{msdnet2018iclr}  and \textit{entropy} \cite{teerapittayanon2016branchynet}-based \textit{pixel-level} confidence estimators, commonly used in multi-exit classification. Our experiments with various architectural configurations indicate that the proposed exit-policy offers a consistently better speed-accuracy trade-off compared to corresponding averaging counterparts (directly generalising from classification-based metrics by averaging per-pixel confidences for each image), with accuracy gains of up to 6.34 pp (1.17~pp on average). %across the spectrum of workloads.
An example of this trade-off is illustrated in Fig. \ref{fig:confmetric}.

\subsection{Qualitative Evaluation}
\label{sec:eval_qualitative}
In this section, we qualitatively analyse some important properties of MESS networks.

\begin{figure}%[h]  
\vspace{0.7cm}
    \centering
	\includegraphics[trim =15mm 35mm 30mm 6mm, clip,width=1\columnwidth]{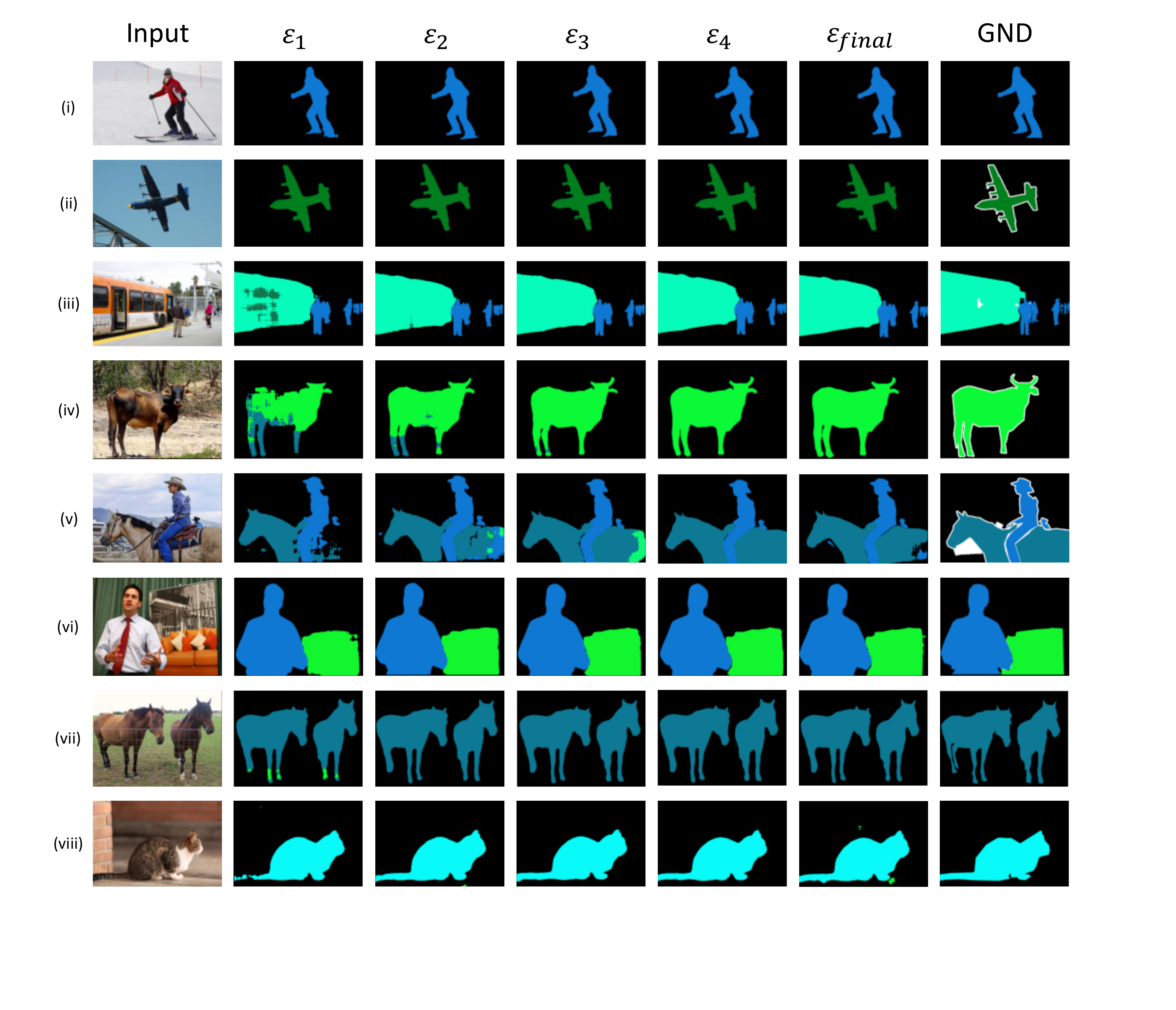}
	\caption{\scriptsize
 Visualisation of per-exit semantic segmentation outputs of MESS ResNet50 for specific samples from MS COCO and PASCAL VOC. The first column represents the input image and last the ground truth labels. Intermediate columns represent the output per exit head.}
	\label{fig:qualitative}
\end{figure}

Initially, Fig.~\ref{fig:qualitative} demonstrates the quality-of-result for progressive segmentation outputs through the exits of a MESS network, for certain samples from MS COCO and PASCAL VOC. Table~\ref{tab:qualitative_acc} lists the respective accuracy for the same set of images, and indicates the output selected by our input-dependent exit policy. 

\begin{figure}%[h]  
\vspace{1cm}
    \centering
	\includegraphics[trim =15mm 50mm 40mm 50mm, clip,width=1.03\columnwidth]{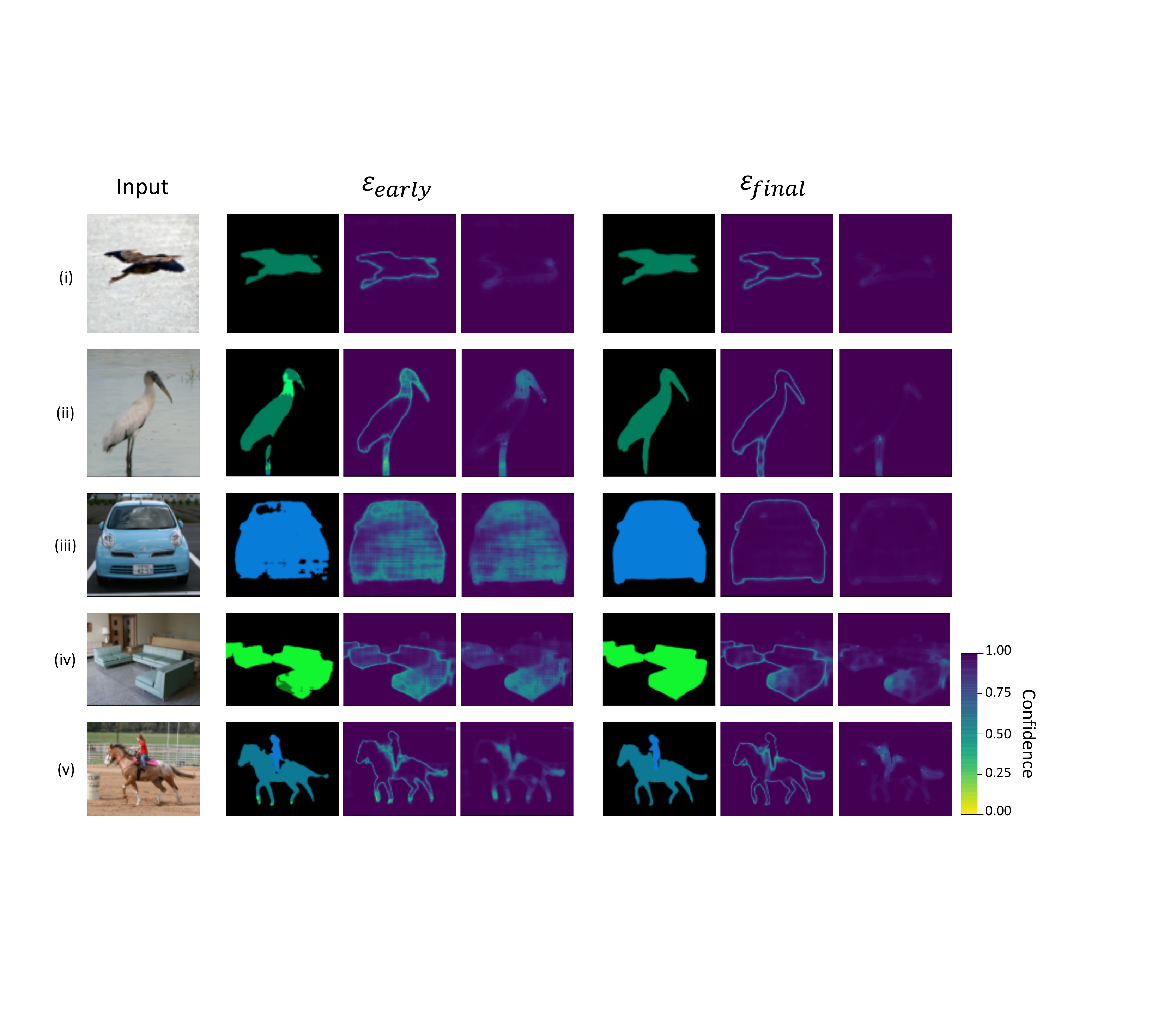}
      \put(-244.5,0){\scriptsize Baseline}
      \put(-195.5,0){\scriptsize Ours}
      \put(-106.5,0){\scriptsize Baseline}
      \put(-58,0){\scriptsize Ours}
      \caption{\scriptsize Per-pixel confidence maps for an early and the final segmentation heads for ResNet50 with and without integrating Eq.~(\ref{eq:final_conf_map}). A solid purple colour is illustrating the best possible result, where the MESS network is most confident about the per pixel labels. With our edge smoothing technique (Section~\ref{sec:policy}) we witness a more pragmatic confidence estimation for predictions along the edges of objects for both exits. This is especially important for early heads of MESS networks, as it helps distinguishing between ``truly" under-confident and edge-rich predictions, leading to higher early-exit rates with respective latency gains.}
	\label{fig:confmaps}
\end{figure}

Subsequently, in Fig.~\ref{fig:confmaps} we illustrate the (per-pixel) confidence heatmap for an early segmentation head and the final exit for certain samples of the same datasets, with and without Eq.~(\ref{eq:final_conf_map}). This demonstrates how our confidence-based mechanism works in the realm of semantic segmentation and the contribution our edge smoothing technique in the confidence of predictions along object edges. Similarly, Table~\ref{tab:qualitative_conf} shows the single (per-image) confidence values for each prediction, obtained both through a baseline and the proposed method.

\begin{figure}%[!b]  
    \vspace{5mm}
    \centering
	\includegraphics[trim =5mm 170mm 48mm 0mm, clip,width=1.\columnwidth]{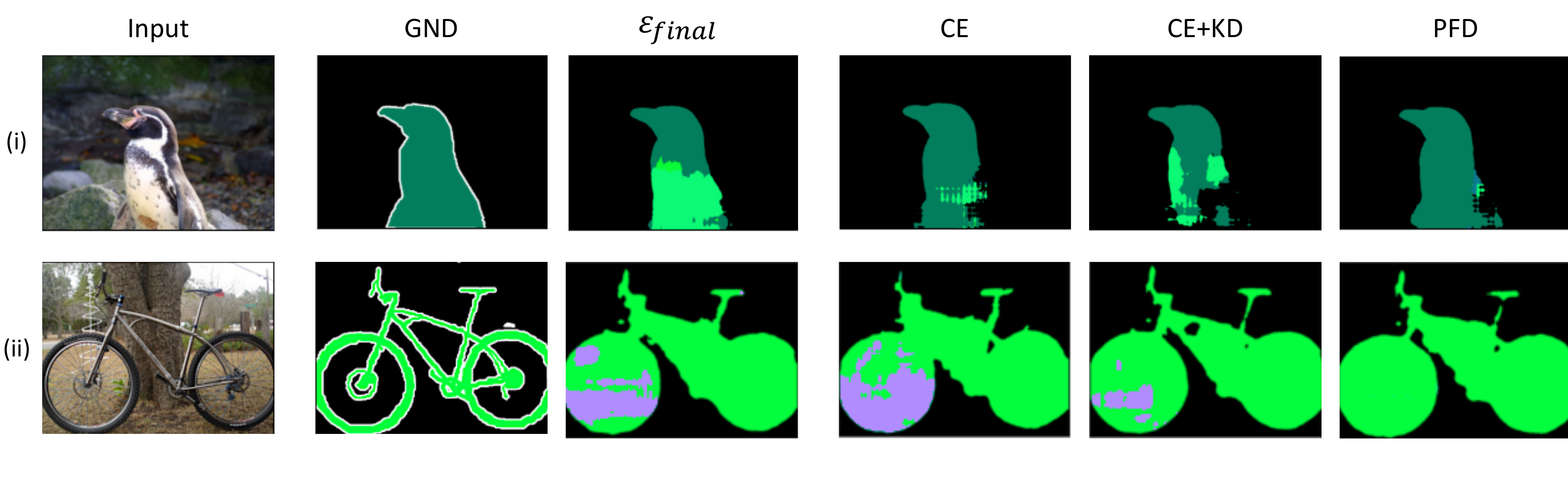}
 \vspace{-7mm}
	\caption{\scriptsize Qualitative examples of semantic segmentation with different distillation schemes on ResNet50. From left to right we see the input image, the ground truth semantic map, the output of the final exit ($\mathcal{E}_{\text{final}}$) and the output of an early exit without distillation (CE), with baseline distillation (CE+KD) and with the proposed positive filtering distillation (PFD) approach. The proposed scheme aims to control the flow of information to early exits in knowledge distillation, ``paying more attention" to pixels that are correctly predicted by the final exit during training, while avoiding to use contradicting CE  and KD reference signals in the remainder of the image. The provided samples illustrate the kind of information that can be learned by means of the PFD scheme, even in the case of the original final exit being incorrect for certain pixels.}
	\label{fig:distil}
 \vspace{5mm}
\end{figure}

Finally, Fig.~\ref{fig:distil} depicts the qualitative difference of semantic map outputs with and without the proposed distillation mechanism (Eq.\ref{eq:loss}) incorporated during training the respective MESS models. The two samples of the figure show clearly the kind of per-pixel prediction errors that our PFD scheme tries to alleviate.

%%%%%%%%%%%%%%%%%%%%%%%

\section{Discussion and Future Work}
\label{sec:eval_limitations}
%\blue{We have shown that MESS networks offer considerable computational gains, both at training and inference time, by leveraging redundancies in the depth-dimension of the DNN, skipping unnecessary computation on demand. In contrast, NAS techniques can attenuate redundancy in more dimensions for efficient model design, yet in an offline manner and at the expense of search time. Nevertheless, we deem the two approaches largely complementary, and future work could combine the two, by applying the MESS methodology on top of a NAS-crafted backbone network. Moreover, softmax-based confidence can be an artificial proxy for measuring a network's uncertainty \cite{hapi2020iccad,adaptive_dnns2021emdl,calibration2017icml}, applicable to classification. Thus, alternative, trainable exit policies and metrics can be a promising avenue of research.}

MESS networks offer considerable computational gains by alleviating redundancies across the depth dimension of the backbone network, skipping unnecessary computation in a difficulty-aware manner. %on demand. % similar to other early-exit approaches~\cite{adaptive_dnns2021emdl}.
In contrast, other efficient model design methodologies, such as NAS \cite{liu2019auto}, can attenuate redundancy in more dimensions (\textit{e.g.}~number of channels and spatial resolution) at the cost of prolonged training, search and inference times. With the two approaches having different benefits and 
capitalising on completely orthogonal directions to obtain efficiency gains (NAS focuses on uniformly eliminating redundancy  \textit{on the backbone}, whereas MESS \textit{enhances a given backbone} with early exits to minimise computational redundancy in an adaptive-inference manner) future work could combine the two, by applying the MESS methodology on top of a NAS-crafted backbone, realising complementary performance gains. Moreover, softmax-based confidence can be an artificial proxy for measuring a network's uncertainty~\cite{hapi2020iccad,adaptive_dnns2021emdl,calibration2017icml}, applicable to classification. Thus, alternative, trainable exit policies and metrics can be a promising avenue of research.

\section{Conclusion}
In this paper, we have presented the concept and realisation of multi-exit semantic segmentation.
% We have proposed MESS networks, 
Applicable to state-of-the-art CNN approaches, MESS models perform efficient semantic segmentation, without sacrificing accuracy.
This is achieved by introducing novel training and early-exiting techniques, tailored for MESS networks. Post-training, our framework can customise the MESS network by searching for the optimal multi-exit configuration (number, placement and architecture of exits) according to the target platform, pushing the limits of efficient deployment.

\begin{table}%[h]
\caption{ \scriptsize Per-exit segmentation accuracy for samples of Fig.~\ref{fig:qualitative} from MS COCO and PASCAL VOC Validation Set. \textit{mIoU} represents the normalised mean IoU per image and \textit{pAcc} the pixel-accuracy (excl. True Positives on \textit{background} class) for each exit head. We denote the selected exit for each sample, determined by the proposed MESS exit policy,  with \textbf{bold} font. Different exits have different confidence thresholds, selected during search, so as to lead to $\leq1$ pp of accuracy degradation.}
\begin{center}
\renewcommand{\arraystretch}{1.0} 
\setlength\tabcolsep{3pt} % default value: 6pt
\resizebox{1\columnwidth}{!}{
\begin{tabular}{ ccc|cc|cc|cc|cc}
 \hline
  \textbf{Sample} & \multicolumn{2}{c}{ $\mathcal{E}_1$ } &  \multicolumn{2}{c}{ $\mathcal{E}_2$ } &  \multicolumn{2}{c}{ $\mathcal{E}_3$ } & \multicolumn{2}{c}{ $\mathcal{E}_4$ } & \multicolumn{2}{c}{ $\mathcal{E}_{final}$ } \\
  & mIoU & pAcc & mIoU & pAcc & mIoU & pAcc & mIoU & pAcc & mIoU & pAcc \\ 
  \hline
    (i) & 94.79\% & 90.70\% &   \textbf{95.91\%} & \textbf{92.69\%} &   95.90\% & 92.67\%&   95.83\% & 92.53\% &   95.82\% & 92.53\% \\
    (ii) & \textbf{92.89\%}  & \textbf{ 86.78\% } & 90.40\%   & 82.12\%   & 91.93\%    & 84.98\%  & 93.32\%  & 87.58\%  &  93.35\% & 87.62\%  \\
    (iii) & 84.61\%  & 82.30\% & 88.68\%  & 92.29\%   & \textbf{88.98\%}  & \textbf{92.77\%}  & 89.16\%   & 93.07\%  & 88.19\%    & 93.07\%  \\
    (iv) &  83.18\% & 68.73\% & 95.67\%  & 91.82\%  & \textbf{98.51\%}   & \textbf{97.58\%}  & 99.03\% & 98.53\%  & 98.97\%   & 98.45 \%  \\
    (v) &  73.06\% & 65.14\%  & 81.11\%  & 82.06\%   & 96.68\%   & 90.01\%  & \textbf{97.64\% } & \textbf{98.31\%}  & 90.78\%   & 90.27\%  \\
    (vi) & 92.47\%  & 92.87\%  & 95.24\% & 95.34\%   & \textbf{95.46\% } & \textbf{95.54\% } & 95.43\% & 95.53\%  & 95.25\%    & 95.34\%  \\
    (vii) &  93.07\%  & 89.88\%   & \textbf{94.49\%}  & \textbf{92.72\% } & 94.59\%    & 92.87\%  & 94.56\%    & 92.85\% & 94.54\% & 92.78\%  \\
    (viii) &  93.98\% & 90.75\% & \textbf{95.58\% }  & \textbf{93.09\%}   & 95.47\%  & 93.12\%  & 92.98\%  & 93.89\%   &  95.31\%   & 92.10\%  \\
    \hline
\end{tabular}
}

\end{center}
\vspace{-1.5em}
\label{tab:qualitative_acc} 
\vspace{0.7cm}
\end{table}

\begin{table}%[h]
\caption{ \scriptsize Per-image confidence values for samples illustrated in Fig.~\ref{fig:confmaps}, reduced from the respective per-pixel confidence maps, through a baseline confidence-averaging approach and the proposed technique (considering the percentage of confident pixels at the output) with and without integrating the semantic edge confidence smoothing of Eq.~(\ref{eq:final_conf_map}). Using the proposed per-image confidence estimation methodology for dense predictions, a better separation is achieved between confident predictions (corresponding to higher quality-of-result outputs) and less confident predictions (prone to semantic errors).
% TODO: Listing Confidence values for samples of fig, using baseline, thresholded and masked approaches.
}
\begin{center}
\renewcommand{\arraystretch}{1.0} 
\setlength\tabcolsep{3pt} % default value: 6pt
{ 
%\resizebox{0.82\columnwidth}{!}{
\begin{tabular}{ cccc|ccc}
 \hline
  \textbf{Sample} & \multicolumn{3}{c}{ $\mathcal{E}_{early}$ } &   \multicolumn{3}{c}{ $\mathcal{E}_{final}$ } \\
  & mean($c^{map}$) & Eq.~(\ref{eq:confidence}) & Eq.~(\ref{eq:confidence}) + (\ref{eq:final_conf_map})  & mean($c^{map}$) & Eq.~(\ref{eq:confidence}) & Eq.~(\ref{eq:confidence}) + (\ref{eq:final_conf_map})  \\
  \hline
    (i)  & 0.990 & 0.975 & 0.999 &   0.992 & 0.977 & 1.000 \\
    (ii) & 0.968 & 0.920 & 0.946 &   0.983 & 0.951 & 0.988 \\
    (iii)& 0.871 & 0.595 & 0.599 &   0.983 & 0.958 & 0.999 \\
    (iv) & 0.905 & 0.725 & 0.734 &   0.958 & 0.874 & 0.918 \\
    (v) & 0.967 & 0.913 & 0.939 &   0.975 & 0.924 & 0.966 \\
    \hline

\end{tabular}
}
%}
\end{center}
\vspace{-1.5em}
\label{tab:qualitative_conf} 
\vspace{1cm}
\end{table}

%%%%%%%%%%%%%%%%%%%%%%%
{\small
\bibliographystyle{ieee_fullname}
\bibliography{main}

\begin{thebibliography}{10}\itemsep=-1pt

\bibitem{embench_2019}
Mario Almeida, Stefanos Laskaridis, Ilias Leontiadis, Stylianos~I. Venieris,
  and Nicholas~D. Lane.
\newblock {EmBench: Quantifying Performance Variations of Deep Neural Networks
  Across Modern Commodity Devices}.
\newblock In {\em The 3rd International Workshop on Deep Learning for Mobile
  Systems and Applications (EMDL)}, 2019.

\bibitem{badrinarayanan2017segnet}
Vijay Badrinarayanan, Alex Kendall, and Roberto Cipolla.
\newblock {SegNet: A Deep Convolutional Encoder-Decoder Architecture for Image
  Segmentation}.
\newblock {\em IEEE Transactions on Pattern Analysis and Machine Intelligence
  (TPAMI)}, 39(12):2481--2495, 2017.

\bibitem{adaptive2017icml}
Tolga Bolukbasi, Joseph Wang, Ofer Dekel, and Venkatesh Saligrama.
\newblock {Adaptive Neural Networks for Efficient Inference}.
\newblock In {\em International Conference on Machine Learning (ICML)}, pages
  527--536, 2017.

\bibitem{chen2018searching}
Liang-Chieh Chen, Maxwell Collins, Yukun Zhu, George Papandreou, Barret Zoph,
  Florian Schroff, Hartwig Adam, and Jon Shlens.
\newblock Searching for efficient multi-scale architectures for dense image
  prediction.
\newblock In {\em Advances in Neural Information Processing Systems (NeurIPS)},
  pages 8699--8710, 2018.

\bibitem{chen2017deeplab}
Liang-Chieh Chen, George Papandreou, Iasonas Kokkinos, Kevin Murphy, and Alan~L
  Yuille.
\newblock {DeepLab: Semantic Image Segmentation with Deep Convolutional Nets,
  Atrous Convolution, and Fully Connected CRFs}.
\newblock {\em IEEE Transactions on Pattern Analysis and Machine Intelligence
  (TPAMI)}, 40(4):834--848, 2017.

\bibitem{chen2017rethinking}
Liang-Chieh Chen, George Papandreou, Florian Schroff, and Hartwig Adam.
\newblock {Rethinking Atrous Convolution for Semantic Image Segmentation}.
\newblock {\em arXiv preprint arXiv:1706.05587}, 2017.

\bibitem{chen2018encoder}
Liang-Chieh Chen, Yukun Zhu, George Papandreou, Florian Schroff, and Hartwig
  Adam.
\newblock {Encoder-Decoder with Atrous Separable Convolution for Semantic Image
  Segmentation}.
\newblock In {\em European Conference on Computer Vision (ECCV)}, pages
  801--818, 2018.

\bibitem{cheng2019leveraging}
Feiyang Cheng, Hong Zhang, Ding Yuan, and Mingui Sun.
\newblock {Leveraging semantic segmentation with learning-based confidence
  measure}.
\newblock {\em Neurocomputing}, 329:21--31, 2019.

\bibitem{deng2009imagenet}
Jia Deng, Wei Dong, Richard Socher, Li-Jia Li, Kai Li, and Li Fei-Fei.
\newblock {ImageNet: A Large-Scale Hierarchical Image Database}.
\newblock In {\em IEEE Conference on Computer Vision and Pattern Recognition
  (CVPR)}, pages 248--255, 2009.

\bibitem{everingham2010pascal}
Mark Everingham, Luc Van~Gool, Christopher~KI Williams, John Winn, and Andrew
  Zisserman.
\newblock {The Pascal Visual Object Classes (VOC) Challenge}.
\newblock {\em International Journal of Computer Vision (IJCV)},
  88(2):303--338, 2010.

\bibitem{nestdnn2018mobicom}
Biyi Fang, Xiao Zeng, and Mi Zhang.
\newblock {NestDNN: Resource-Aware Multi-Tenant On-Device Deep Learning for
  Continuous Mobile Vision}.
\newblock In {\em Annual International Conference on Mobile Computing and
  Networking (MobiCom)}, page 115–127, 2018.

\bibitem{figurnov2017spatially}
Michael Figurnov, Maxwell~D Collins, Yukun Zhu, Li Zhang, Jonathan Huang,
  Dmitry Vetrov, and Ruslan Salakhutdinov.
\newblock {Spatially Adaptive Computation Time for Residual Networks}.
\newblock In {\em IEEE Conference on Computer Vision and Pattern Recognition
  (CVPR)}, pages 1039--1048, 2017.

\bibitem{gao2019dynamic}
Xitong Gao, Yiren Zhao, Łukasz Dudziak, Robert Mullins, and Cheng zhong Xu.
\newblock {Dynamic Channel Pruning: Feature Boosting and Suppression}.
\newblock In {\em International Conference on Learning Representations (ICLR)},
  2019.

\bibitem{ghiasi2016laplacian}
Golnaz Ghiasi and Charless~C Fowlkes.
\newblock Laplacian pyramid reconstruction and refinement for semantic
  segmentation.
\newblock In {\em European Conference on Computer Vision (ECCV)}, pages
  519--534. Springer, 2016.

\bibitem{ghosh2019understanding}
Swarnendu Ghosh, Nibaran Das, Ishita Das, and Ujjwal Maulik.
\newblock {Understanding Deep Learning Techniques for Image Segmentation}.
\newblock {\em ACM Computing Surveys (CSUR)}, 52(4):1--35, 2019.

\bibitem{calibration2017icml}
Chuan Guo, Geoff Pleiss, Yu Sun, and Kilian~Q. Weinberger.
\newblock {On Calibration of Modern Neural Networks}.
\newblock In {\em International Conference on Machine Learning (ICML)}, 2017.

\bibitem{hariharan2011semantic}
Bharath Hariharan, Pablo Arbel{\'a}ez, Lubomir Bourdev, Subhransu Maji, and
  Jitendra Malik.
\newblock {Semantic Contours from Inverse Detectors}.
\newblock In {\em International Conference on Computer Vision (ICCV)}, pages
  991--998, 2011.

\bibitem{he2016deep}
Kaiming He, Xiangyu Zhang, Shaoqing Ren, and Jian Sun.
\newblock {Deep Residual Learning for Image Recognition}.
\newblock In {\em IEEE Conference on Computer Vision and Pattern Recognition
  (CVPR)}, pages 770--778, 2016.

\bibitem{hinton2015distilling}
Geoffrey Hinton, Oriol Vinyals, and Jeff Dean.
\newblock {Distilling the Knowledge in a Neural Network}.
\newblock In {\em NeurIPS 2014 Deep Learning Workshop}, 2014.

\bibitem{gatingnns2019neurips}
Weizhe Hua, Yuan Zhou, Christopher~M De~Sa, Zhiru Zhang, and G.~Edward Suh.
\newblock {Channel Gating Neural Networks}.
\newblock In {\em Advances in Neural Information Processing Systems (NeurIPS)},
  pages 1886--1896, 2019.

\bibitem{msdnet2018iclr}
Gao Huang, Danlu Chen, Tianhong Li, Felix Wu, Laurens van~der Maaten, and
  Kilian Weinberger.
\newblock {Multi-Scale Dense Networks for Resource Efficient Image
  Classification}.
\newblock In {\em International Conference on Learning Representations (ICLR)},
  2018.

\bibitem{huang2017densely}
Gao Huang, Zhuang Liu, Laurens Van Der~Maaten, and Kilian~Q Weinberger.
\newblock {Densely Connected Convolutional Networks}.
\newblock In {\em IEEE Conference on Computer Vision and Pattern Recognition
  (CVPR)}, pages 4700--4708, 2017.

\bibitem{ai_benchmark_2019}
Andrey Ignatov, Radu Timofte, Andrei Kulik, Seungsoo Yang, Ke Wang, Felix Baum,
  Max Wu, Lirong Xu, and Luc Van~Gool.
\newblock {AI Benchmark: All About Deep Learning on Smartphones in 2019}.
\newblock In {\em International Conference on Computer Vision (ICCV)
  Workshops}, 2019.

\bibitem{reda2020mm}
Junguang Jiang, Ximei Wang, Mingsheng Long, and Jianmin Wang.
\newblock {Resource Efficient Domain Adaptation}.
\newblock In {\em ACM International Conference on Multimedia (MM)}, 2020.

\bibitem{sdn2019icml}
Yigitcan Kaya, Sanghyun Hong, and Tudor Dumitras.
\newblock {Shallow-Deep Networks: Understanding and Mitigating Network
  Overthinking}.
\newblock In {\em International Conference on Machine Learning (ICML)}, 2019.

\bibitem{adaptive_dnns2021emdl}
Stefanos Laskaridis, Alexandros Kouris, and Nicholas~D. Lane.
\newblock {Adaptive Inference through Early-Exit Networks: Design, Challenges
  and Directions}.
\newblock In {\em Proceedings of the 5th International Workshop on Embedded and
  Mobile Deep Learning (EMDL)}, page 1–6, 2021.

\bibitem{spinn2020mobicom}
Stefanos Laskaridis, Stylianos~I. Venieris, Mario Almeida, Ilias Leontiadis,
  and Nicholas~D. Lane.
\newblock {SPINN}: {S}ynergistic {P}rogressive {I}nference of {N}eural
  {N}etworks over {D}evice and {C}loud.
\newblock In {\em Annual International Conference on Mobile Computing and
  Networking (MobiCom)}. ACM, 2020.

\bibitem{hapi2020iccad}
Stefanos Laskaridis, Stylianos~I. Venieris, Hyeji Kim, and Nicholas~D. Lane.
\newblock {HAPI}: {H}ardware-{A}ware {P}rogressive {I}nference.
\newblock In {\em International Conference on Computer-Aided Design (ICCAD)},
  2020.

\bibitem{persephonee2021hotmobile}
Ilias Leontiadis, Stefanos Laskaridis, Stylianos~I. Venieris, and Nicholas~D.
  Lane.
\newblock {It's Always Personal: Using Early Exits for Efficient On-Device CNN
  Personalisation}.
\newblock In {\em Proceedings of the 22nd International Workshop on Mobile
  Computing Systems and Applications (HotMobile)}, 2021.

\bibitem{ee_training2019iccv}
Hao Li, Hong Zhang, Xiaojuan Qi, Ruigang Yang, and Gao Huang.
\newblock {Improved Techniques for Training Adaptive Deep Networks}.
\newblock In {\em IEEE International Conference on Computer Vision (ICCV)},
  2019.

\bibitem{li2017not}
Xiaoxiao Li, Ziwei Liu, Ping Luo, Chen Change~Loy, and Xiaoou Tang.
\newblock {Not All Pixels Are Equal: Difficulty-aware Semantic Segmentation via
  Deep Layer Cascade}.
\newblock In {\em IEEE Conference on Computer Vision and Pattern Recognition
  (CVPR)}, pages 3193--3202, 2017.

\bibitem{li2020learning}
Yanwei Li, Lin Song, Yukang Chen, Zeming Li, Xiangyu Zhang, Xingang Wang, and
  Jian Sun.
\newblock {Learning Dynamic Routing for Semantic Segmentation}.
\newblock In {\em IEEE/CVF Conference on Computer Vision and Pattern
  Recognition (CVPR)}, pages 8553--8562, 2020.

\bibitem{lin2017refinenet}
Guosheng Lin, Anton Milan, Chunhua Shen, and Ian Reid.
\newblock {Refinenet: Multi-Path Refinement Networks for High-Resolution
  Semantic Segmentation}.
\newblock In {\em IEEE Conference on Computer Vision and Pattern Recognition
  (CVPR)}, pages 1925--1934, 2017.

\bibitem{runtime_pruning2017neurips}
Ji Lin, Yongming Rao, Jiwen Lu, and Jie Zhou.
\newblock {Runtime Neural Pruning}.
\newblock In {\em Advances in Neural Information Processing Systems (NeurIPS)},
  pages 2181--2191, 2017.

\bibitem{lin2014microsoft}
Tsung-Yi Lin, Michael Maire, Serge Belongie, James Hays, Pietro Perona, Deva
  Ramanan, Piotr Doll{\'a}r, and C~Lawrence Zitnick.
\newblock {Microsoft COCO: Common Objects in Context}.
\newblock In {\em European Conference on Computer Vision (ECCV)}, pages
  740--755, 2014.

\bibitem{liu2019auto}
Chenxi Liu, Liang-Chieh Chen, Florian Schroff, Hartwig Adam, Wei Hua, Alan~L
  Yuille, and Li Fei-Fei.
\newblock {Auto-DeepLab: Hierarchical Neural Architecture Search for Semantic
  Image Segmentation}.
\newblock In {\em IEEE Conference on Computer Vision and Pattern Recognition
  (CVPR)}, pages 82--92, 2019.

\bibitem{edge_ar2019mobicom}
Luyang Liu, Hongyu Li, and Marco Gruteser.
\newblock {Edge Assisted Real-Time Object Detection for Mobile Augmented
  Reality}.
\newblock In {\em Annual International Conference on Mobile Computing and
  Networking (MobiCom)}, 2019.

\bibitem{liu2019structured}
Yifan Liu, Ke Chen, Chris Liu, Zengchang Qin, Zhenbo Luo, and Jingdong Wang.
\newblock {Structured Knowledge Distillation for Semantic Segmentation}.
\newblock In {\em IEEE Conference on Computer Vision and Pattern Recognition
  (CVPR)}, 2019.

\bibitem{long2015fully}
Jonathan Long, Evan Shelhamer, and Trevor Darrell.
\newblock {Fully Convolutional Networks for Semantic Segmentation}.
\newblock In {\em IEEE Conference on Computer Vision and Pattern Recognition
  (CVPR)}, pages 3431--3440, 2015.

\bibitem{luan2019msd}
Yunteng Luan, Hanyu Zhao, Zhi Yang, and Yafei Dai.
\newblock {MSD: Multi-Self-Distillation Learning via Multi-classifiers within
  Deep Neural Networks}.
\newblock {\em arXiv:1911.09418}, 2019.

\bibitem{luc2016semantic}
Pauline Luc, Camille Couprie, Soumith Chintala, and Jakob Verbeek.
\newblock {Semantic Segmentation using Adversarial Networks}.
\newblock In {\em NIPSW on Adversarial Training}, 2016.

\bibitem{mccormac2017semanticfusion}
John McCormac, Ankur Handa, Andrew Davison, and Stefan Leutenegger.
\newblock {SemanticFusion: Dense 3D Semantic Mapping with Convolutional Neural
  Networks}.
\newblock In {\em 2017 IEEE International Conference on Robotics and Automation
  (ICRA)}, pages 4628--4635. IEEE, 2017.

\bibitem{mehta2018espnet}
Sachin Mehta, Mohammad Rastegari, Anat Caspi, Linda Shapiro, and Hannaneh
  Hajishirzi.
\newblock {ESPNet: Efficient Spatial Pyramid of Dilated Convolutions for
  Semantic Segmentation}.
\newblock In {\em European Conference on Computer Vision (ECCV)}, pages
  552--568, 2018.

\bibitem{nekrasov2019fast}
Vladimir Nekrasov, Hao Chen, Chunhua Shen, and Ian Reid.
\newblock {Fast Neural Architecture Search of Compact Semantic Segmentation
  Models via Auxiliary Cells}.
\newblock In {\em IEEE Conference on Computer Vision and Pattern Recognition
  (CVPR)}, pages 9126--9135, 2019.

\bibitem{noh2015learning}
Hyeonwoo Noh, Seunghoon Hong, and Bohyung Han.
\newblock {Learning Deconvolution Network for Semantic Segmentation}.
\newblock In {\em IEEE International Conference on Computer Vision (ICCV)},
  pages 1520--1528, 2015.

\bibitem{nvidia_video_conf}
NVIDIA.
\newblock {NVIDIA Maxine - Cloud-AI Video-Streaming Platform}.
\newblock \url{https://developer.nvidia.com/maxine}, 2020.
\newblock {[Retrieved: \today]}.

\bibitem{peng2017large}
Chao Peng, Xiangyu Zhang, Gang Yu, Guiming Luo, and Jian Sun.
\newblock {Large Kernel Matters--Improve Semantic Segmentation by Global
  Convolutional Network}.
\newblock In {\em IEEE Conference on Computer Vision and Pattern Recognition
  (CVPR)}, pages 4353--4361, 2017.

\bibitem{phuong2019distillation}
Mary Phuong and Christoph~H Lampert.
\newblock {Distillation-based Training for Multi-Exit Architectures}.
\newblock In {\em IEEE International Conference on Computer Vision (ICCV)},
  pages 1355--1364, 2019.

\bibitem{ronneberger2015u}
Olaf Ronneberger, Philipp Fischer, and Thomas Brox.
\newblock {U-Net: Convolutional Networks for Biomedical Image Segmentation}.
\newblock In {\em International Conference on Medical Image Computing and
  Computer-Assisted Intervention}, pages 234--241. Springer, 2015.

\bibitem{sandler2018mobilenetv2}
Mark Sandler, Andrew Howard, Menglong Zhu, Andrey Zhmoginov, and Liang-Chieh
  Chen.
\newblock {MobileNetV2: Inverted Residuals and Linear Bottlenecks}.
\newblock In {\em IEEE Conference on Computer Vision and Pattern Recognition
  (CVPR)}, pages 4510--4520, 2018.

\bibitem{Siam_2018_CVPR_Workshops}
Mennatullah Siam, Mostafa Gamal, Moemen Abdel-Razek, Senthil Yogamani, Martin
  Jagersand, and Hong Zhang.
\newblock {A Comparative Study of Real-Time Semantic Segmentation for
  Autonomous Driving}.
\newblock In {\em Conference on Computer Vision and Pattern Recognition (CVPR)
  Workshops}, 2018.

\bibitem{43022}
Christian Szegedy, Wei Liu, Yangqing Jia, Pierre Sermanet, Scott Reed, Dragomir
  Anguelov, Dumitru Erhan, Vincent Vanhoucke, and Andrew Rabinovich.
\newblock {Going Deeper with Convolutions}.
\newblock In {\em IEEE Conference on Computer Vision and Pattern Recognition
  (CVPR)}, 2015.

\bibitem{teerapittayanon2016branchynet}
Surat Teerapittayanon, Bradley McDanel, and Hsiang-Tsung Kung.
\newblock {BranchyNet: Fast Inference via Early Exiting from Deep Neural
  Networks}.
\newblock In {\em 2016 23rd International Conference on Pattern Recognition
  (ICPR)}, pages 2464--2469. IEEE, 2016.

\bibitem{adaptive_layer2018eccv}
Andreas Veit and Serge Belongie.
\newblock {Convolutional Networks with Adaptive Inference Graphs}.
\newblock In {\em European Conference on Computer Vision (ECCV)}, pages 3--18,
  2018.

\bibitem{vu2019advent}
Tuan-Hung Vu, Himalaya Jain, Maxime Bucher, Matthieu Cord, and Patrick
  P{\'e}rez.
\newblock {ADVENT: Adversarial Entropy Minimization for Domain Adaptation in
  Semantic Segmentation}.
\newblock In {\em IEEE Conference on Computer Vision and Pattern Recognition
  (CVPR)}, pages 2517--2526, 2019.

\bibitem{wang2018skipnet}
Xin Wang, Fisher Yu, Zi-Yi Dou, Trevor Darrell, and Joseph~E Gonzalez.
\newblock {SkipNet: Learning Dynamic Routing in Convolutional Networks}.
\newblock In {\em European Conference on Computer Vision (ECCV)}, pages
  409--424, 2018.

\bibitem{wang2020dynamic}
Yulong Wang, Xiaolu Zhang, Xiaolin Hu, Bo Zhang, and Hang Su.
\newblock {Dynamic Network Pruning with Interpretable Layerwise Channel
  Selection}.
\newblock In {\em AAAI Conference on Artificial Intelligence (AAAI)}, pages
  6299--6306, 2020.

\bibitem{wu2019fastfcn}
Huikai Wu, Junge Zhang, Kaiqi Huang, Kongming Liang, and Yu Yizhou.
\newblock {FastFCN: Rethinking Dilated Convolution in the Backbone for Semantic
  Segmentation}.
\newblock In {\em arXiv preprint arXiv:1903.11816}, 2019.

\bibitem{wu2019demystifying}
Yanzhao Wu, Ling Liu, Juhyun Bae, Ka-Ho Chow, Arun Iyengar, Calton Pu, Wenqi
  Wei, Lei Yu, and Qi Zhang.
\newblock {Demystifying Learning Rate Policies for High Accuracy Training of
  Deep Neural Networks}.
\newblock In {\em IEEE International Conference on Big Data (Big Data)}, pages
  1971--1980, 2019.

\bibitem{wu2018blockdrop}
Zuxuan Wu, Tushar Nagarajan, Abhishek Kumar, Steven Rennie, Larry~S Davis,
  Kristen Grauman, and Rogerio Feris.
\newblock {BlockDrop: Dynamic Inference Paths in Residual Networks}.
\newblock In {\em IEEE Conference on Computer Vision and Pattern Recognition
  (CVPR)}, pages 8817--8826, 2018.

\bibitem{deebert2020acl}
Ji Xin, Raphael Tang, Jaejun Lee, Yaoliang Yu, and Jimmy Lin.
\newblock {{D}ee{BERT}: Dynamic Early Exiting for Accelerating {BERT}
  Inference}.
\newblock In {\em 58th Annual Meeting of the Association for Computational
  Linguistics (ACL)}, pages 2246--2251, 2020.

\bibitem{xing2020early}
Qunliang Xing, Mai Xu, Tianyi Li, and Zhenyu Guan.
\newblock {Early Exit Or Not: Resource-Efficient Blind Quality Enhancement for
  Compressed Images}.
\newblock In {\em European Conference on Computer Vision (ECCV)}, 2020.

\bibitem{xu2017end}
Huazhe Xu, Yang Gao, Fisher Yu, and Trevor Darrell.
\newblock End-to-end learning of driving models from large-scale video
  datasets.
\newblock In {\em IEEE Conference on Computer Vision and Pattern Recognition
  (CVPR)}, pages 2174--2182, 2017.

\bibitem{balancedsparse2019aaai}
Zhuliang Yao, Shijie Cao, Wencong Xiao, Chen Zhang, and Lanshun Nie.
\newblock {Balanced Sparsity for Efficient DNN Inference on GPU}.
\newblock In {\em AAAI Conference on Artificial Intelligence (AAAI)},
  volume~33, pages 5676--5683, 2019.

\bibitem{heimdall2020mobicom}
Juheon Yi and Youngki Lee.
\newblock {Heimdall: Mobile GPU Coordination Platform for Augmented Reality
  Applications}.
\newblock In {\em {Annual International Conference on Mobile Computing and
  Networking (MobiCom)}}, 2020.

\bibitem{yu2018bisenet}
Changqian Yu, Jingbo Wang, Chao Peng, Changxin Gao, Gang Yu, and Nong Sang.
\newblock {BiSeNet: Bilateral Segmentation Network for Real-Time Semantic
  Segmentation}.
\newblock In {\em European Conference on Computer Vision (ECCV)}, pages
  325--341, 2018.

\bibitem{yu2016mutli}
Fisher Yu and Vladlen Koltun.
\newblock {Multi-Scale Context Aggregation by Dilated Convolutions}.
\newblock In {\em International Conference on Learning Representations (ICLR)},
  2016.

\bibitem{yu2017dilated}
Fisher Yu, Vladlen Koltun, and Thomas Funkhouser.
\newblock {Dilated Residual Networks}.
\newblock In {\em IEEE Conference on Computer Vision and Pattern Recognition
  (CVPR)}, pages 472--480, 2017.

\bibitem{yuan2019s2dnas}
Zhihang Yuan, Bingzhe Wu, Zheng Liang, Shiwan Zhao, Weichen Bi, and Guangyu
  Sun.
\newblock {S2DNAS: Transforming Static CNN Model for Dynamic Inference via
  Neural Architecture Search}.
\newblock In {\em European Conference on Computer Vision (ECCV)}, 2020.

\bibitem{telepresence2020eccv}
E. Zakharov, Aleksei Ivakhnenko, Aliaksandra Shysheya, and V. Lempitsky.
\newblock {Fast Bi-layer Neural Synthesis of One-Shot Realistic Head Avatars}.
\newblock In {\em European Conference on Computer Vision (ECCV)}, 2020.

\bibitem{cardiacsegnas2020iccad}
Dewen Zeng, Weiwen Jiang, Tianchen Wang, Xiaowei Xu, Haiyun Yuan, Meiping
  Huang, Jian Zhuang, Jingtong Hu, and Yiyu Shi.
\newblock {Towards Cardiac Intervention Assistance: Hardware-aware Neural
  Architecture Exploration for Real-Time 3D Cardiac Cine MRI Segmentation}.
\newblock In {\em ACM/IEEE International Conference on Computer-Aided Design
  (ICCAD)}, 2020.

\bibitem{zhang2019be}
Linfeng Zhang, Jiebo Song, Anni Gao, Jingwei Chen, Chenglong Bao, and Kaisheng
  Ma.
\newblock {Be your Own Teacher: Improve the Performance of Convolutional Neural
  Networks via Self Distillation}.
\newblock In {\em IEEE International Conference on Computer Vision (ICCV)},
  2019.

\bibitem{scan2019neurips}
Linfeng Zhang, Zhanhong Tan, Jiebo Song, Jingwei Chen, Chenglong Bao, and
  Kaisheng Ma.
\newblock {SCAN: A Scalable Neural Networks Framework Towards Compact and
  Efficient Models}.
\newblock In {\em Advances in Neural Information Processing Systems (NeurIPS)},
  2019.

\bibitem{zhao2018icnet}
Hengshuang Zhao, Xiaojuan Qi, Xiaoyong Shen, Jianping Shi, and Jiaya Jia.
\newblock {ICNet for Real-Time Semantic Segmentation on High-Resolution
  Images}.
\newblock In {\em European Conference on Computer Vision (ECCV)}, pages
  405--420, 2018.

\bibitem{zhao2017pyramid}
Hengshuang Zhao, Jianping Shi, Xiaojuan Qi, Xiaogang Wang, and Jiaya Jia.
\newblock {Pyramid Scene Parsing Network}.
\newblock In {\em IEEE Conference on Computer Vision and Pattern Recognition
  (CVPR)}, pages 2881--2890, 2017.

\bibitem{zhou2019edge}
Zhi Zhou, Xu Chen, En Li, Liekang Zeng, Ke Luo, and Junshan Zhang.
\newblock {Edge Intelligence: Paving the Last Mile of Artificial Intelligence
  with Edge Computing}.
\newblock {\em Proceedings of the IEEE}, 107(8):1738--1762, 2019.

\end{thebibliography}
}

%%%%%%%%%%%%%%%%%%%%%%%

\renewcommand{\thesection}{A}
\newpage
\section{Supplemental Material}
%\vspace{-7mm}
%\etocsettocstyle{\rule{\linewidth}{\tocrulewidth}\vskip0.5\baselineskip}{\rule{\linewidth}{\tocrulewidth}}
%\localtableofcontents
%\vspace{-4mm}

In the supplementary material of our paper, we provide further details about the experimental setup of our work, discussing the examined datasets and baselines, as well as the adopted training and inference protocols. We also provide additional quantitative results, relevant to the architectural choices of MESS networks.

\subsection{Experimental Configuration}
\label{sec:apdx_exp_setup}

\subsubsection{Datasets.} In this paper we evaluate MESS networks on the following datasets:
\vspace{1mm}

\noindent  \textbf{\textit{MS COCO:}} MS COCO~\cite{lin2014microsoft} forms one of the largest datasets for dense scene understanding tasks. Thereby, it acts  as common ground for pre-training semantic segmentation models across domains. Following common practice for semantic segmentation, we consider only the 20 semantic classes of PASCAL VOC~\cite{everingham2010pascal} (plus a \textit{background} class), and discard any training images that consist solely of background pixels. This results in 92.5k training and 5k validation images. We set crop size ($b_R$) to 520$\times$520.

\vspace{1mm}
\noindent \textbf{\textit{PASCAL VOC:}} PASCAL VOC~\cite{everingham2010pascal} comprises the most broadly used benchmark for semantic segmentation. It includes 20 foreground object classes (plus a \textit{background} class). The original dataset consists of 1464 training and 1449 validation images. Following common practise we adopt the augmented training set provided by \cite{hariharan2011semantic}, resulting in 10.5k training images. For PASCAL VOC, $b_R$ is also set to 520$\times$520.

%\noindent \textbf{Cityscapes(?):} Cityscapes forms another widely used semantic segmentation benchmark, focusing on urban scenes and autonomous driving scenarios. The original dataset contains 30 classes, but the commonly used benchmark only focuses on 19 classes (plus a \textit{background} class). The dataset contains 2975 and 500 finely annotated (1024$\times$2048) images for training and validation respectively. Another set of 20K coarse annotations also exits, but is not used in this work. We set the $b_R$ for Cityscapes to 768$\times$1536.

\subsubsection{Baselines.}
\vspace{-2mm}
To compare our work's performance with the state-of-the-art, we evaluate against the following approaches: 

\vspace{2mm}
\noindent {Single-Exit Segmentation Backbones:} \vspace{-1mm}
\begin{itemize}[leftmargin=0.5cm]
    \setlength\itemsep{-0mm}
    \item \textbf{DRN}: Dilated Residual Networks~\cite{yu2017dilated} approach for re-using classification pre-trained CNNs as backbones for semantic segmentation, by avoiding loss of spatial information. We use an FCN head at the end. 
    \item \textbf{DLBV3}: DeepLabV3~\cite{yu2017dilated,chen2018encoder}, one of the leading approaches in semantic segmentation, employing Atrous Spatial Pyramid Pooling (ASPP). 
    \item \textbf{segMBNetV2}: The lightweight MobileNetV2 segmentor presented in \cite{sandler2018mobilenetv2} with an FCN head. 
\end{itemize}
\noindent {Mutli-Exit Segmentation SOTA:} \vspace{-1mm}
\begin{itemize}[leftmargin=0.5cm]
    \setlength\itemsep{-0mm}
    \item \textbf{LC}: The early-exit segmentation work Deep Layer Cascade~\cite{li2017not}. \textbf{LC} proposes a pixel-wise adaptive propagation in early-exit segmentation networks, with confident pixel-level predictions exiting early. 
\end{itemize}

\noindent {NAS Segmentation SOTA:} \vspace{-1mm}
\begin{itemize}[leftmargin=0.5cm]
    \setlength\itemsep{0mm}
        \item \textbf{AutoDLB}: The NAS-based segmentation approach Auto-DeepLab~\cite{liu2019auto}, employing a differential formulation for hierarchical NAS, leading to high search efficiency. We target the Auto-DeepLab-M variant.
\end{itemize}

\newpage
\noindent {Multi-Exit Network Training:} \vspace{-0.5mm}
\begin{itemize}[leftmargin=0.5cm]
    \setlength\itemsep{0mm}
    \item \textbf{E2E}\footnote{Adapted for segmentation by incorporating dense predictions.\label{note2}}: The conventional end-to-end training method for early-exit classification networks, introduced by MSDNet~\cite{msdnet2018iclr} and BranchyNet~\cite{teerapittayanon2016branchynet}. 
    \item \textbf{Frozen}$^{\ref{note2}}$: The conventional frozen-backbone training method for early-exit classification networks, proposed by SDN~\cite{sdn2019icml} and HAPI~\cite{hapi2020iccad}. 
\end{itemize}

\noindent {Distillation-based Training:} \vspace{-0.5mm}
\begin{itemize}[leftmargin=0.5cm]
\setlength\itemsep{0mm}
    \item \textbf{KD}$^{\ref{note2}}$: The originally proposed knowledge distillation technique of \cite{hinton2015distilling}.
    \item \textbf{SelfDistill}$^{\ref{note2}}$: The popular self-distillation approach for early-exit classification networks, utilised in~\cite{scan2019neurips,phuong2019distillation,zhang2019be}. 
\end{itemize}

\vspace{-2mm}
\subsection{Training Protocol}
\vspace{-0mm}
MESS instances are built on top of existing segmentation networks, spanning across the workload spectrum in the literature, \textit{i.e.}~from the computationally heavy \cite{chen2017rethinking} to the lightweight \cite{sandler2018mobilenetv2}. Through MESS, SOTA networks can be \textit{further optimised for deployment efficiency}, demonstrating complementary performance gains by saving computation on easier samples. A key characteristic of the proposed two-stage MESS training scheme, is that it effectively preserves the accuracy of the final (baseline) exit, while boosting the attainable results to earlier segmentation exits. This is achieved by bringing together elements from both the end-to-end and frozen-backbone training approaches (Sec. \ref{sec:pre-training}). Thus, we consider the employed training scheme decoupled from the attainable comparative results, as long as both the baseline and the corresponding MESS instances share the same training procedure. As such, in order to preserve simplicity in this work, we use a straightforward training scheme, shared across all networks and datasets, and refrain from exotic data augmentation, bootstrapping and multi-stage pre-training schemes that can be found in accuracy-centric approaches. 

\vspace{1mm}
\noindent \textit{\textbf{Hyperpameters:}} All  MESS and baseline models are optimised using SGD, starting from ImageNet~\cite{deng2009imagenet} pre-trained backbones. The initial learning rate is set to  $lr_0$=0.02 and  \textit{poly} lr-schedule ($lr_0 \cdot (1-\frac{iter}{total\ iter})^{pow}$) \cite{wu2019demystifying} with $pow$=0.9 is employed. Training runs over 60k iterations in all datasets. Momentum is set to 0.9 and weight decay to $10^{-4}$. We re-scale all images to a dataset-dependent base resolution $b_R$. During training we conduct the following \textit{data augmentation} techniques: random re-scaling by 0.5$\times$ to 2.0$\times$, random cropping (size: 0.9$\times$) and random horizontal flipping ($p=0.5$). For Knowledge Distillation, we experimentally set $\alpha$ to 0.5. For the overprovisioned network, we experimentally found $N$=6 to provide a good balance between search space granularity and size, for the examined backbones.

\subsection{Inference Process}
\vspace{-1mm}
The main optimisation objective of this work is \textit{deployment efficiency}. This renders impractical many popular inference strategies that are broadly utilised in  the literature when optimising solely for accuracy, as they incur prohibitive workload overheads. As such, in this work, we refrain from the use of ensembles, multi-grid and multi-scale inference, image flipping, etc. Instead, in the context of this work, both MESS and baseline networks employ a straightforward single-pass inference across all inputs.

\subsection{Comparison with Uniform Exit Architectures}
Compared to a direct adoption of classification-based approaches that employ a uniform exit architecture across the depth of the backbone~\cite{msdnet2018iclr,teerapittayanon2016branchynet,sdn2019icml,hapi2020iccad}, MESS networks provide a significantly improved performance-accuracy trade-off that pushes the limits of efficient execution for semantic segmentation, by providing a highly customisable architectural configuration space for early exits, searched though our framework.

\begin{figure}[H]
    \centering
    \begin{subfigure}[t]{0.44\linewidth}
    \centering
	\includegraphics[trim =1mm 0mm 42mm 0mm, clip,width=1\linewidth]{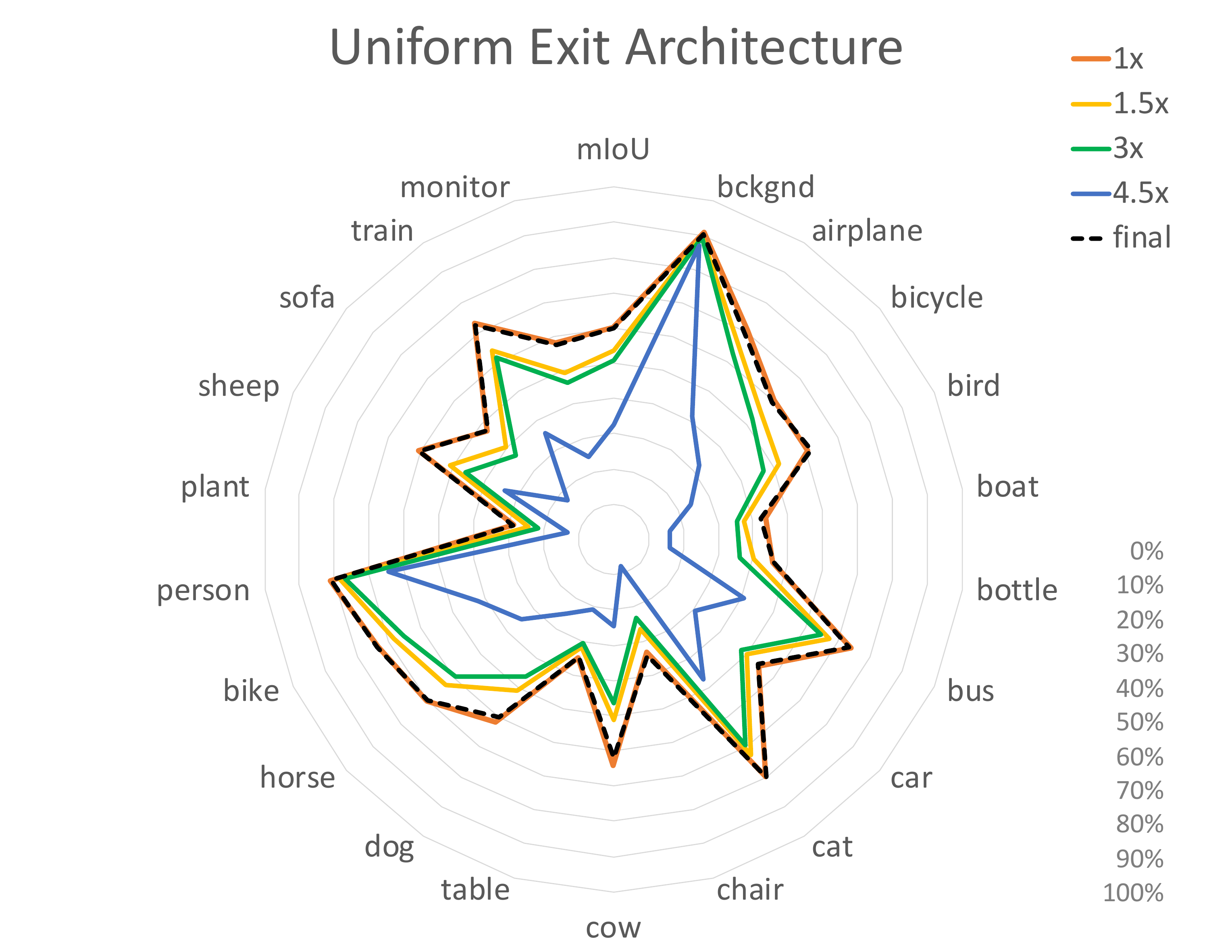}
	%\caption{DRAFT: Configurable Exit Architecture}
	%\caption{}
	\label{fig:spider_uni}
    \end{subfigure}%
    ~ 
    \begin{subfigure}[t]{0.477\linewidth}
    \centering
	\includegraphics[trim =22mm 0mm 2mm 2mm, clip,width=1\linewidth]{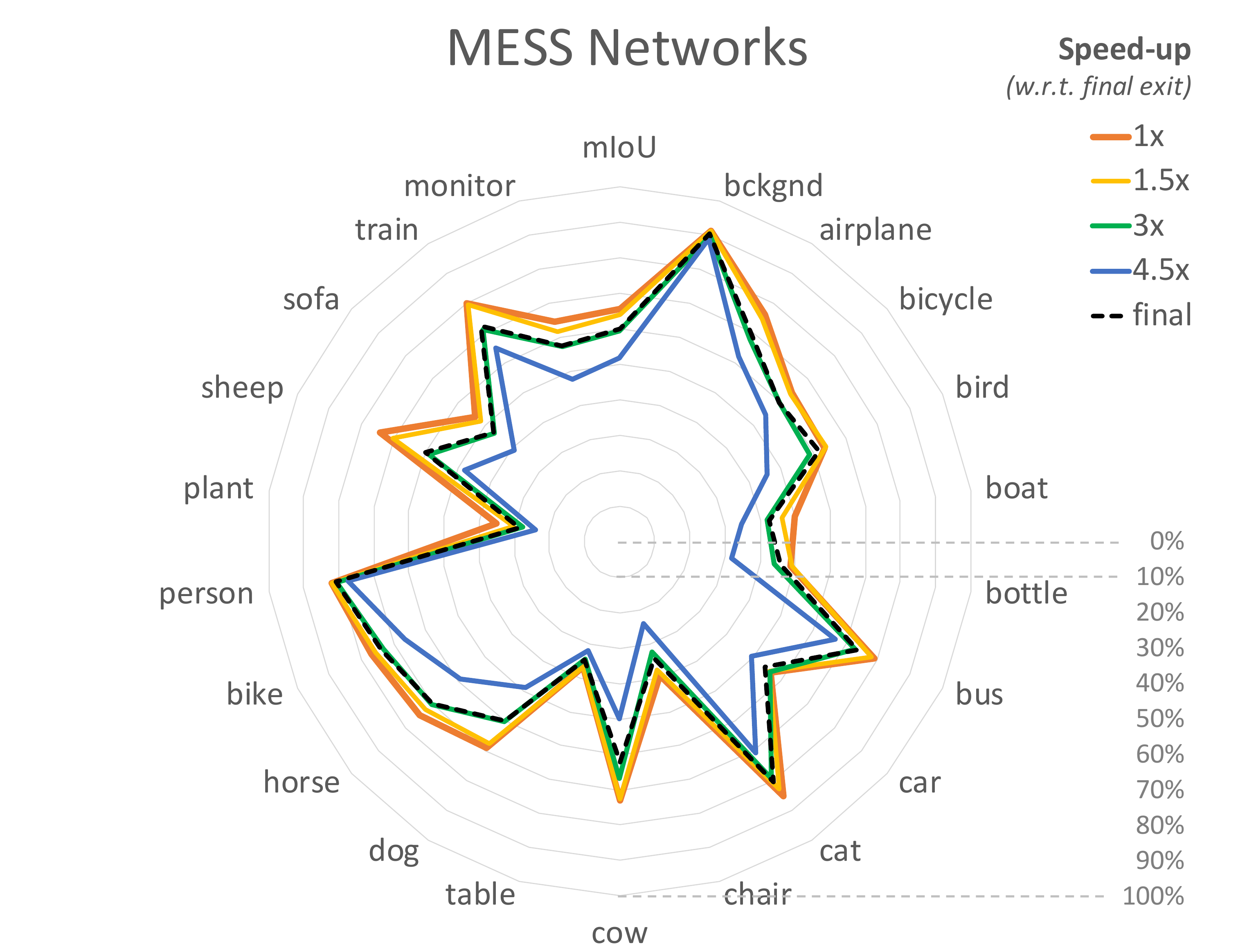}
	%\caption{DRAFT: Configurable Exit Architecture}
    %\caption{}
	\label{fig:spider_mess}
    \end{subfigure}
    \put(-300,4){\small (a)}
	\put(-155,4){\small (b)}
	\vspace{-3mm}
    \caption{\small Per-label IoU on MS COCO Validation Set of (a) uniform architecture early-exit networks and (b) MESS networks configured for different latency goals (expressed as speed-ups with respect to the final exit), for ResNet50. Each of the concentric spider graphs defines a distribution of exit rates over different heads, based on the confidence threshold determined by our search. The graphs have value both when studied in isolation, so as to monitor the behaviour of early exits across labels, and in comparison, to understand what our principled design approach offers.
    }
    \label{fig:spider}
%    \vspace{2mm}
\end{figure}

To demonstrate this, Fig.~\ref{fig:spider} illustrates the mean and per-label accuracy of multi-exit network instances, customised in view of requirements ranging between $1\times$ and $4.5\times$ lower latency compared to the original backbone. On the left side we depict baseline networks using a \textit{uniform exit architecture} of FCN heads across all candidate exit points. In contrast, on the right side we examine MESS networks incorporating \textit{tailored early-exit architectures} from the proposed search space. %The results indicate that MESS networks can boost accuracy (mIoU) up to 19 pp (11 pp on average) compared to their uniform exit-architecture counterparts, meeting the same latency requirements.
The results indicate significant accuracy (mIoU) gains by exploiting the proposed head, ranging up to \textbf{19.3 pp} (10.9 pp on avg.), across the examined latency budgets. This demonstrates that it is essential to re-design the segmentation heads for multi-exiting scenarios. Repeating the same experiment, but optimising for latency under an accuracy constraint, MESS reduces FLOPs by up to \textbf{3}$\bm{\times}$ (2.4$\times$ on avg.), across varying accuracy targets.

\vspace{5mm}

\end{document}